\documentclass{article}

\usepackage{microtype}
\usepackage{graphicx}
\usepackage{subfigure}
\usepackage{booktabs} % for professional tables
\newcommand{\otoprule}{\midrule[\heavyrulewidth]}
\usepackage{multirow}
\usepackage{hyperref}

\usepackage[accepted]{icml2023}

\usepackage{amsmath}
\usepackage{amssymb}
\usepackage{mathtools}
\usepackage{amsthm}

\usepackage[capitalize,noabbrev]{cleveref}

\theoremstyle{plain}

\theoremstyle{definition}

\theoremstyle{remark}

\usepackage[textsize=tiny]{todonotes}
\DeclareMathOperator*{\argmin}{arg\,min}

\icmltitlerunning{Taylorformer: Probabilistic Modelling for Random Processes including Time Series}

\begin{document}

\twocolumn[
\icmltitle{Taylorformer: Probabilistic Modelling for  \\
           Random Processes including Time Series}

% It is OKAY to include author information, even for blind
% submissions: the style file will automatically remove it for you
% unless you've provided the [accepted] option to the icml2023
% package.

% List of affiliations: The first argument should be a (short)
% identifier you will use later to specify author affiliations
% Academic affiliations should list Department, University, City, Region, Country
% Industry affiliations should list Company, City, Region, Country

% You can specify symbols, otherwise they are numbered in order.
% Ideally, you should not use this facility. Affiliations will be numbered
% in order of appearance and this is the preferred way.
\icmlsetsymbol{equal}{*}

\begin{icmlauthorlist}
\icmlauthor{Omer Nivron}{equal,cs}
\icmlauthor{Raghul Parthipan}{equal,cs,bas}
\icmlauthor{Damon J. Wischik}{cs}
%\icmlauthor{}{sch}
%\icmlauthor{}{sch}
\end{icmlauthorlist}

\icmlaffiliation{cs}{Department of Computer Science and Technology, University of Cambridge, UK}
\icmlaffiliation{bas}{British Antarctic Survey, Cambridge, UK}

\icmlcorrespondingauthor{Omer Nivron}{on234@cam.ac.uk}
\icmlcorrespondingauthor{Raghul Parthipan}{rp542@cam.ac.uk}

% You may provide any keywords that you
% find helpful for describing your paper; these are used to populate
% the "keywords" metadata in the PDF but will not be shown in the document
\icmlkeywords{Machine Learning, ICML}

\vskip 0.3in
]

% this must go after the closing bracket ] following \twocolumn[ ...

% This command actually creates the footnote in the first column
% listing the affiliations and the copyright notice.
% The command takes one argument, which is text to display at the start of the footnote.
% The \icmlEqualContribution command is standard text for equal contribution.
% Remove it (just {}) if you do not need this facility.

%\printAffiliationsAndNotice{}  % leave blank if no need to mention equal contribution
\printAffiliationsAndNotice{\icmlEqualContribution} % otherwise use the standard text.

\begin{abstract}
We propose the Taylorformer for random processes such as time series. Its two key components are: 1) the LocalTaylor wrapper which adapts Taylor approximations (used in dynamical systems) for use in neural network-based probabilistic models, and 2) the MHA-X attention block which makes predictions in a way inspired by how Gaussian Processes' mean predictions are linear smoothings of contextual data. Taylorformer outperforms the state-of-the-art in terms of log-likelihood on 5/6 classic Neural Process tasks such as meta-learning 1D functions, and has at least a 14\% MSE improvement on forecasting tasks, including electricity, oil temperatures and exchange rates. Taylorformer approximates a consistent stochastic process and provides uncertainty-aware predictions. Our code is provided in the supplementary material.

\end{abstract}

\section{Introduction}

% *what is new in our approach and why do we think it will work*

% Attention-based models such as GPT \cite{Liu2018} XXADD CITATIONSXX have shown impressive results on natural language tasks, where the goal is to predict sequences of discrete values. These models are built on the standard multi-head attention introduced by \citet{Vaswani2017}. However, the scientific community is still working out what adaptions are needed to get similarly impressive results for continuous problems (such as time-series or spatial processes), where the goal is to predict real-valued targets on a continuum. For example, the Transformer Neural Process (TNP) \citep{pmlr-v162-nguyen22b} uses standard attention mechanisms to model continuous problems in an autoregressive manner, and the batch-generating Autoformer \citep{Wu2021AutoformerDT} is tailored for time-series forecasting, incorporating a decomposition block which extracts different temporal trends. 

Stochastic processes are appealing because they provide uncertainty along with point predictions, making them useful for modelling random systems and for decision-making. Limitations of conventional techniques like Gaussian Processes (GP) include high computational costs and difficulties in specifying kernels. Notable efforts to combine the benefits of stochastic processes and neural networks have included leveraging attention, such as the Attentive Neural Process \cite{Kim2019AttentiveNP}, a batch-generation model, and the state-of-the-art Transformer Neural Process (TNP) \cite{pmlr-v162-nguyen22b}, an autoregressive model. Both models are built on the standard multi-head attention introduced by \citet{Vaswani2017}. 

Models based on the standard multi-head attention, such as GPT \cite{Liu2018}, have shown impressive results on natural language tasks, where the goal is to predict sequences of discrete values. However, the scientific community is still working out what adaptations are needed to get similarly impressive results for continuous problems (such as time-series or spatial processes), where the goal is to predict real-valued targets on a continuum. 

% and the Autoformer \citep{Wu2021AutoformerDT}, tailored for time-series forecasting, incorporates a decomposition block which extracts different temporal trends to model in a  batch-generation manner. 

% Omer 16-05 - I don't understand line 20 labelled points 
% Problem setup we have to say what alpha beta are
% It's a problem setup and starts with 'Our goal' ..should change? 

\paragraph{Problem setup.} Our goal is to model the distribution of a set of unobserved points (target), $\mathbf{Y}_T$, at a specified set of indices, $\mathbf{X}_T$, given a set of observed points and their associated indices (context), $\{\mathbf{X}_C,\mathbf{Y}_C\}$. Here, $\mathbf{Y}_T \in \mathbb{R}^{n_T \times \alpha}, \mathbf{X}_T \in \mathbb{R}^{n_T \times \beta}, \mathbf{Y}_C \in \mathbb{R}^{n_C \times \alpha}$ and $\mathbf{X}_C \in \mathbb{R}^{n_C \times \beta}$, where $n_C$ and $n_T$ are the numbers of points in the context and target sets respectively. The context or target set may be permuted arbitrarily --- they need not be ordered.

% The current adaptions of attention often draw inspiration from existing techniques, such as the idea of decomposing time-series [********]. Our work is driven by the questions: \textit{What other canonical modelling approaches for continuous processes are known to be useful? And how can we use them to enhance machine learning models?} Based on these insights, we propose the Taylorformer and show that it achieves state-of-the-art performance.

% Omer 16-05 - line 31 with the smoothing is cryptic 
% Let's think about "Our contributions" one is a contribution but doesnt
% it include already 2, 3, 4?

\paragraph{Our Contributions.} (1) We propose the Taylorformer, a probabilistic model for continuous processes. It produces predictions and associated uncertainty for interpolation and extrapolation settings and irregularly sampled data. It approximates a consistent stochastic process \cite{Garnelo2018NeuralP} and builds on insights derived from two well-known approaches for modelling continuous processes: Taylor series and Gaussian Processes (GPs). (2) We introduce the LocalTaylor wrapper.  Taylor series can be used for local approximations of functions, such as in dynamical systems. But they are only useful under certain conditions. LocalTaylor uses a neural network to learn how and when to use the information provided by a Taylor approximation. (3) We create the MHA-X block to incorporate an inductive bias from GPs (specifically, how the mean prediction is a linear smoothing of contextual values). (4) We introduce a masking procedure for attention units in the Taylorformer to allow training the ability to predict points at arbitrary locations.

\begin{figure*}[th]
  \includegraphics[width=0.8\textwidth]{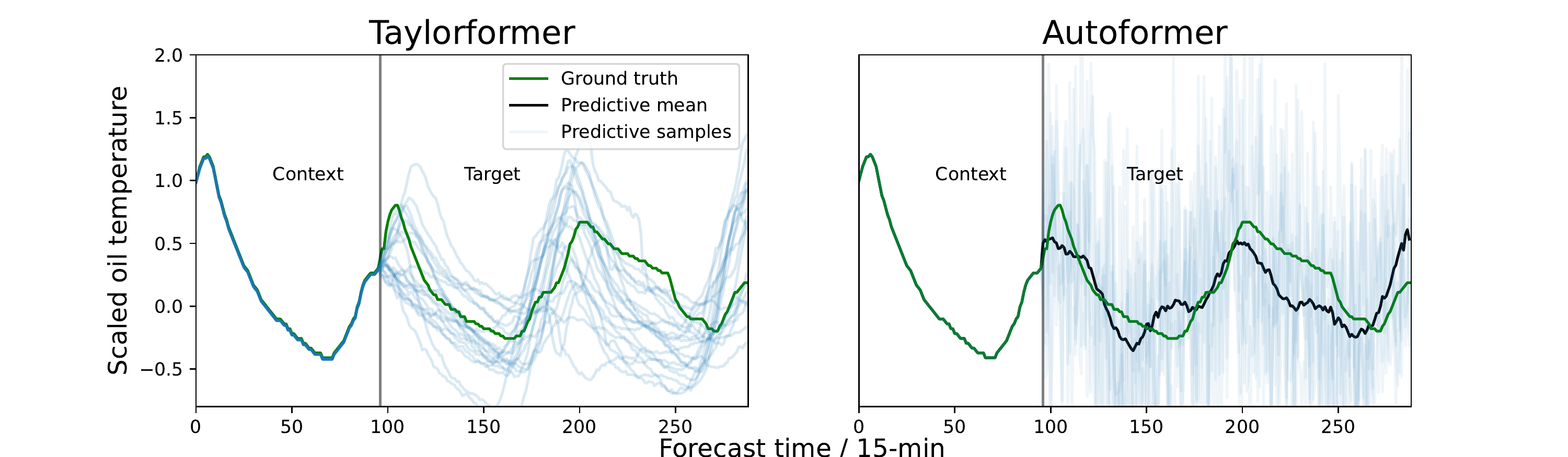}
  \centering
  \caption{Taylorformer (left) can generate higher quality samples on the ETT \cite{Wu2021AutoformerDT} dataset 
  than the state-of-the-art Autoformer \cite{Wu2021AutoformerDT} (right). This is representative of the general difference between these models at generation time. The task is to predict the next 48 hours (192 target points) given a 24-hour window (96 context points).}
  \label{fig:samples_ett}
\end{figure*}

This work focuses on building better models for stochastic processes, so we do not focus on computational cost or inference efficiency.

We demonstrate Taylorformer's performance on several tasks in the Neural Process (NP) and time-series forecasting literature. Together, these evaluate the following i) both mean predictions and modelled uncertainty, ii) in both interpolation and extrapolation settings and iii) consistency, where the context/target sets can either be ordered or take arbitrary permutations. Taylorformer has a better log-likelihood/mean-squared-error than state-of-the-art models on 17/18 tasks, including image completion (2D regression) and electricity consumption. The Taylorformer shows 14-18\%, 21-34\% and 95-99\% reductions in MSE compared to the next best model for forecasting transformers' oil temperatures,  electricity consumption, and exchange rates, respectively. Figure \ref{fig:samples_ett} shows samples from the Taylorformer and the Autoformer for the ETT dataset \cite{Wu2021AutoformerDT}. 
% The samples of the Autoformer are very noisy due its global standard deviation.

% with 21-34\%, 95-99\% and 14-18\% reductions in MSE compared to the closest model for the ETT, exchange and electricity tasks, respectively

% \par
% For the probabilistic modeller, our goal is to maximise the likelihood 
% \[P(Y_{(c+1):m_{j}}|[\textbf{X}_{c}, \textbf{X}_{T}, \textbf{Y}_{c}]) \].

%Let's introduce mathematical notation to help us clarify these five components. We concentrate on a concrete dataset for the explanation. We are trying to forecast the daily temperatures ($^{\circ} C$) in a specific location for the next $m$ days given the information from days $1,2, \dots, c$. We will denote the random variable representing daily temperature at day $t$ as $Y_{t}$ and its readout as $y_{t} \in \mathbf{R}$. We will assume that $Y_{t}$ is normally distributed with parameters $\mu_{t}$ and $\sigma_{t}$ and our goal is to predict these parameters. We call the first $c$ pairs of $(t, y_t)$ values context points, and the subsequent $m$ pairs of $(t, y_t)$ we call target points. The total sequence length is $N = c+m$. We use superscript to indicate layers. When we use the expression $i:j$ in subscript we mean all elements between index $i$ and index $j$ included. Furthermore, we will use square brackets to indicate a concatenation procedure.     
%Note that the sequence pairs do not have to follow a chronological ordering --- this will depend on the application.   

\section{Related Work}

% ***0. There are two communities who work on this sort of thing: NP, and Attention Modellers. 
% They seem to have different interests! We think there's a single project here, and
% our model gets inspiration from both backgrounds.

% ***1. Interest in consistency --- in proper probabilistic Random Process models
% (which are inherently consistent).
% GP people write in the language of probability, whereas we've presented it as prediction.
% We just did that for simplicity; our transformer is of course a probabilistic model.

% ***2. What are the other questions that people ask about time series modelling? --- they're not 
% interested in forecasting, they're interested in efficiency.

There has been a separation in the literature between those that use neural networks to model random processes (the Neural Process family) and those that model time-series. We believe there is a shared project and are motivated by both.
% \vspace{-1.74em}
\paragraph{Neural Processes and Consistency.}

The initial members of the Neural Process family (the Conditional Neural Process, CNP, \cite{Garnelo2018ConditionalNP} and the Neural Process, NP, \cite{Garnelo2018NeuralP}) set out to create a neural network version of Gaussian Processes (GPs), to combine the benefits of both. GPs are probabilistic models which specify distributions over functions. They can be used to perform regression and provide uncertainty estimates. Prior knowledge can be incorporated through the specification of a parametric kernel function, and they can adapt to new observations. One of their desirable properties is that a GP is a consistent stochastic process (roughly, it does not matter what order you present the data in); however, they can be computationally intensive to use, and selecting appropriate kernels is a challenging task. 
 % Omer 16-05 - the following para. does not flow well it's stuck
 
One consequence of consistency is target equivariance which means that for a given probability model with a likelihood function, $p_\theta$, for a given context set $\{\mathbf{X}_C,\mathbf{Y}_C\}$ and target set $\{\mathbf{X}_T,\mathbf{Y}_T\}$
\begin{equation}
    p_\theta(\mathbf{Y}_{\pi(T)}\,|\,\mathbf{Y}_C,\mathbf{X}_C,\mathbf{X}_{\pi(T)}) = p_\theta(\mathbf{Y}_T\,|\,\mathbf{Y}_C,\mathbf{X}_C,\mathbf{X}_T)
    \label{eq:target_equi}
\end{equation}
where $\pi(T)$ is a permutation of the target set. Another consequence is context invariance, which means,
\begin{equation}
    p_\theta(\mathbf{Y}_T\,|\,\mathbf{Y}_C,\mathbf{X}_C,\mathbf{X}_T)
=
p_\theta(\mathbf{Y}_T\,|\,\mathbf{Y}_{\pi(C)},\mathbf{X}_{\pi(C)},\mathbf{X}_T)
\label{eq:context in}
\end{equation}
where $\pi(C)$ is a permutation of the context set. 

% Omer 16-05 - adaptations vs adaptions?
% Omer 16-05 - line 61-63 don't flow
% Omer 16-05 - it's unclear why we make a big deal of consistency, should we start with that? 
% Omer 16-05 - I think this section will work better after our approach

The TNP \cite{pmlr-v162-nguyen22b} is a state-of-the-art autoregressive Neural Process based on the transformer architecture, with adaptations, such as a new mask, for continuous processes. Context invariance is enforced via the architecture, but they rely on a shuffling approach to encourage target equivariance through the training algorithm. \citet{pmlr-v162-nguyen22b} define the desired target-equivariant model by means of its likelihood $\tilde{p}_\theta$, $\tilde{p}_\theta(\mathbf{Y}_{T}\,|\,\mathbf{Y}_C,\mathbf{X}_C,\mathbf{X}_T)
= \mathbb{E}_\pi
[p_\theta(\mathbf{Y}_{\pi(T)}\,|\,\mathbf{Y}_C,\mathbf{X}_C,\mathbf{X}_{\pi(T)})]$ where $p_\theta$ is the likelihood of the base TNP model. The Expectation is approximated using a Monte Carlo average over randomly sampled permutations during training. Inspired by the TNP, we use a similar shuffling procedure for training. 
% This is equivariant as [**********]

The initial Neural Process work (CNP, NP) enforced consistency through constraints to the architectures, but such constraints resulted in the underfitting of data. Advances included using attention in a way that maintains consistency \cite{Kim2019AttentiveNP}, using convolutional architectures \cite{gordon2019convolutional} and using hierarchical latent variables to capture global and local latent information separately \cite{wang2020doubly}. The TNP outperformed all these by trading an increase in flexibility for the loss of architecture-enforced consistency.

\paragraph{Forecasting.}

There have been various improvements to standard Attention layers in the forecasting literature. Two state-of-the-art examples are Informer \cite{Zhou2020InformerBE} and Autoformer \cite{Wu2021AutoformerDT}. Autoformer proposes a replacement for the classic multi-head attention block by using an auto-correlation mechanism, and the Informer offers a more efficient way to choose the most relevant keys for specific queries using ProbSparse attention.

One key difference between Taylorformer and Autoformer/Informer is that we model the targets autoregressively, while the latter uses a batch-generation approach (one-shot generation). Batch-generation approaches model the targets as conditionally independent given the context, analogous to CNPs \cite{Garnelo2018ConditionalNP}. The conditional independence assumption is restrictive, and allowing conditional dependencies between targets can improve results, as seen by how the NP \cite{Garnelo2018NeuralP} improves on the CNP. 

% Moreover, Autoformer, like us, suggests a different treatment for covariates and target points. Whereas, Informer treats both covariates and targets similarly. 

% One-shot generation models are inherently consistent (as defined in \eqref{consistent}) but only for a fixed length pre-defined by the model architecture. Another advantage of one-shot generation is its reduced computational demands when used in generation.   
Much other work focuses on making attention mechanisms more efficient \cite{NEURIPS2020_c8512d14, Wang2020LinformerSW, pmlr-v119-katharopoulos20a, Kitaev2020ReformerTE,Choromanski2020RethinkingAW}. And there is another strand combining both accuracy and efficiency \cite{Wu2021AutoformerDT, Zhou2020InformerBE, LI2019EnhancingTL}. Many are batch-generation models, so they are more efficient at forecast time than autoregressive ones since all the targets can be produced in one shot. As noted earlier, though, the gap that we aim to fill is how to make a better attention model for the data, putting efficiency aside.

\section{Our approach: Taylorformer}

\begin{figure*}[t!]
    \centering
   \includegraphics[scale=0.65,trim={3.55cm 13cm 0 5cm},clip]{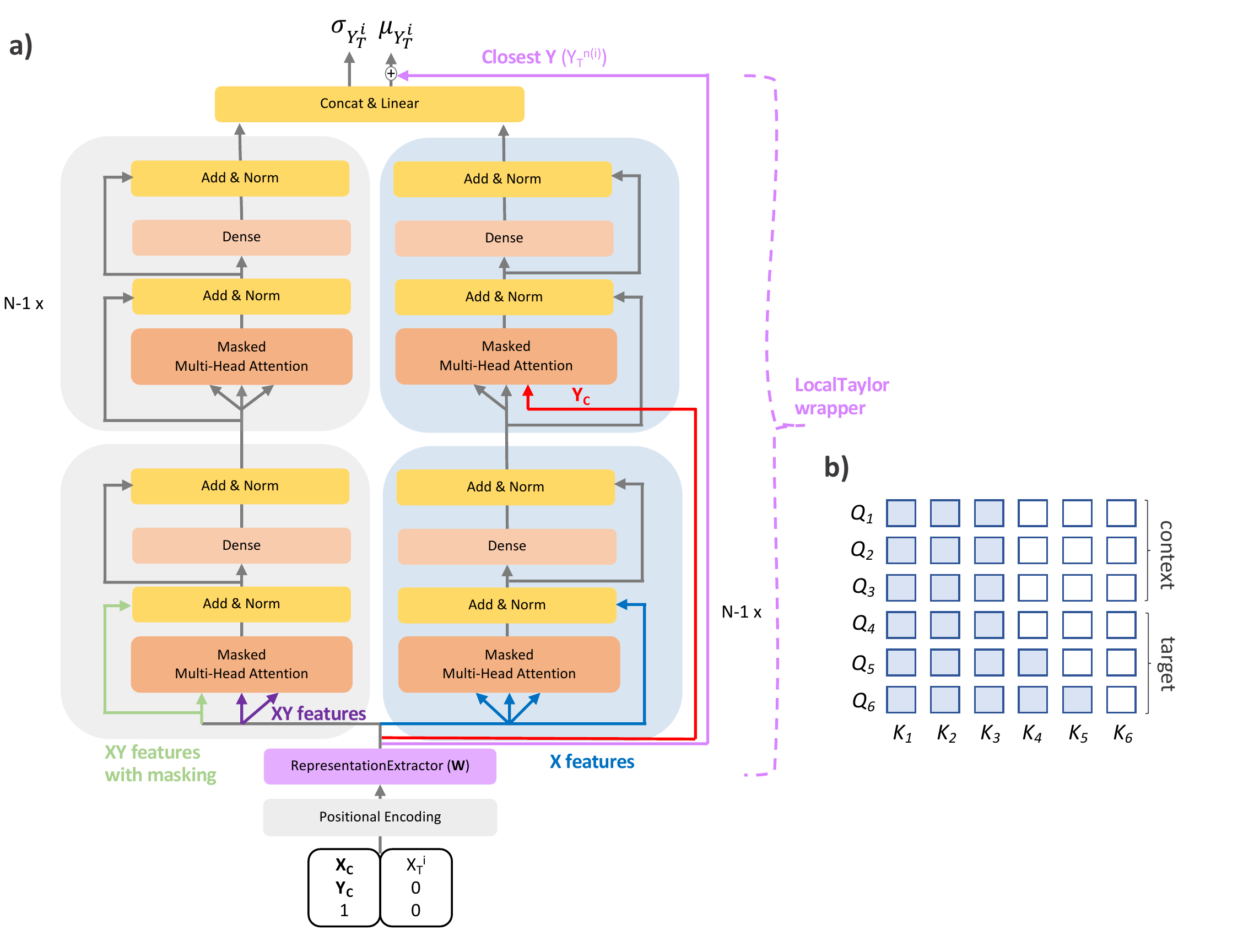}
    \caption{a) Taylorformer architecture corresponding to equation \eqref{eq:our_approach}. LocalTaylor is a wrapper around the central neural network, MHA-X-Net. The channels on the right-hand side are MHA-X. The ones on the left are MHA-XY. The noted features are shown in the following equations: XY features with masking, eq. \eqref{eq:query_xy}, XY features, eq. \eqref{eq:value_xy}, and X features, eq. \eqref{eq:query_x}. b) Example mask for $n_C = 3 $ and $n_T = 3$. Each token can attend to other shaded tokens in its row.}
    \label{fig:architecture}
\end{figure*}

% \includegraphics[]{figures/atp_figure.pdf}

% \begin{figure}
%     \centering
%    \includegraphics[scale=0.7,trim={-2cm 9cm 0 1.2cm},clip]{figures/atp_figure.pdf}
%     \caption{Taylorformer architecture. LocalTaylor is a wrapper around the central neural network, MHA-X-Net. The channels on the right-hand side are MHA-X. The ones on the left are MHA-XY. The noted features are shown in the following equations: XY features with masking, eq. \eqref{eq:query_xy}, XY features, eq. \eqref{eq:value_xy}, and X features, eq. \eqref{eq:query_x}.}
%     \label{fig:architecture}
% \end{figure}

% \begin{figure}
%   \includegraphics[width=\columnwidth]{figures/nll_curves_intro.pdf}
%   \centering
%   \caption{Sample figure caption.}
% \end{figure}

% Omer 16-05 - there is no description of the i superscript ?
% Omer 16-05 - I think we need a signpost before the scary equation number 3.

Taylorformer is an autoregressive probabilistic model. The likelihood assigned to $\mathbf{Y}_T$ given $\mathbf{X}_T,\mathbf{Y}_C,\mathbf{X}_C$ is decomposed in the typical autoregressive manner
\begin{multline}
    p(\mathbf{Y}_T|\mathbf{X}_T,\mathbf{Y}_C,\mathbf{X}_C) = p(Y_T^1|X_T^1,\mathbf{Y}_C,\mathbf{X}_C) \times \\\prod_{i=2}^{n_T}p(Y_T^i|X_T^i,\mathbf{Y}_T^{1:i},\mathbf{X}_T^{1:i},\mathbf{Y}_C,\mathbf{X}_C)
\end{multline}
where $\mathbf{Y}_T \in \mathbb{R}^{n_T \times \alpha}, \mathbf{X}_T \in \mathbb{R}^{n_T \times \beta}, \mathbf{Y}_C \in \mathbb{R}^{n_C \times \alpha},\mathbf{X}_C \in \mathbb{R}^{n_C \times \beta}$ and $Y_T^i \in \mathbb{R}^{\alpha}$. The superscript notation $1:i$ means all indices from $1$ to $i-1$.  

% **** I would like to say something here about "this equation reflects the full contribution" , it is currently underplayed.  *** 

The following equation reflects our full contribution. We model $Y_T^i$ given $X_T^i$ and a set of $\{\mathbf{X},\mathbf{Y}\}$ information (which will contain context points and already-predicted target points) as follows, illustrated here using $\{\mathbf{X}_C,\mathbf{Y}_C\}$
% \begin{equation}
%   % Y_T^i = \text{TaylorBlock(NeuralNet},\mathbf{X}_C,X_T^i,\mathbf{Y}_C;p,q) + \text{Linear}(\text{NeuralNet}(\mathbf{W},\mathbf{X}_C,X_T^i,\mathbf{Y}_C)) \cdot Z_i 
%   Y_T^i = ....
% \end{equation}
\begin{multline}
    Y_T^i = \operatorname{LocalTaylor}(\operatorname{MHA-X-Net},X_T^i,\mathbf{X}_C,\mathbf{Y}_C;p,q) + \\
    \operatorname{Linear}(\operatorname{MHA-X-Net}(\mathbf{W},X_T^i,\mathbf{X}_C,\mathbf{Y}_C)) \cdot Z_i
    \label{eq:our_approach}
\end{multline}
where LocalTaylor is detailed in section \ref{method:taylor}, and $p$ and $q$ are its hyperparameters. It is a wrapper around the neural network MHA-X-Net and uses approximations based on the Taylor series to augment predictions. Part of the LocalTaylor is the creation of the features $\mathbf{W}$ (equation \eqref{eq:rep_extractor}). The MHA-X-Net is a neural network composed of standard multi-head attention units (referred to here as MHA-XY) and our new MHA-X block (section \ref{ref:method_mhax}). $Z^i \sim N(0,1)$. The model architecture is shown in Figure \ref{fig:architecture}a. In section \ref{method:training}, we explain how the model is trained.

% Omer 16-05 - I don't fully understand the figure caption 

\subsection{LocalTaylor}
\label{method:taylor}

% We first motivate why Taylor approximations can be useful for us. We then discuss why they are not suited in their vanilla form to the goal of making both short and long-range predictions, nor for noisy data. We then introduce the LocalTaylor wrapper to address these issues.
% [**** adjust signposting*****]
% Omer 16-05 - I think we need to say what is f, y, t etc..
% Omer 16-05 - line 105 requires a citation 
% Omer 16-05 - is y_a, y evalauted at a or is it an index? 
% Omer 16-05 - what's the purpose of lines 108-109? 
Taylor polynomials (curtailed Taylor series) are useful for making local predictions for functions, in low-noise settings, based on information at neighbouring points. Such approaches are useful in machine learning contexts too. The usefulness of a Taylor approximation is especially clear in the modelling of dynamical systems \cite{brenowitz2018prognostic,gagne2020machine,liu2022hierarchical,parthipan2022using, Chen2018NeuralOD, Ruthotto2018DeepNN, pmlr-v80-lu18d}, where one goal is to emulate the fixed-time-step evolution of a system that evolves based on an ordinary differential equation (ODE) of the form $    \frac{dy}{dt} = f(y)$ where an approach based on modelling residuals
\begin{equation}
    y_{a+\delta t} = y_{a} + \delta t \operatorname{NeuralNet}(y_a)
    \label{eq:res_trick}
\end{equation}
often works far better than modelling the target outright, as in 
\begin{equation}
    y_{a+\delta t} = \operatorname{NeuralNet}(y_a)
\end{equation}
where $\operatorname{NeuralNet}$ is a machine learning function. In this example, equation \eqref{eq:res_trick} has a clear parallel with the first order Taylor approximation of $y_{a+\delta t}$ about $y_a$ which is 
     $y_{a+\delta t} = y_{a} + \delta t \left.\frac{dy}{dt}\right\vert_{t=a}$. The neural network in equation \eqref{eq:res_trick} can be interpreted as a term to model all higher-order terms in the Taylor expansion. 

% The results section shows [***** edit/adapt this paragraph ********] This is illustrated in Figures 1a-c, where we show the validation loss curves for models of the form  \eqref{eq:res_trick} and () for three different dynamical systems (Lorenz 96, Brusselator and the Kuramoto-Sivashinsky). In all cases, using the Taylor-like inductive bias results in better models. 
% Omer 16-05 - I don;t understand the last lines of this paragraph.
Taylor approximations are not obviously suited for making long-range predictions (i.e., where $\delta t$ in equation \eqref{eq:res_trick} is large) nor when the data is too noisy. The naive fix for long-range predictions would be to include higher-order terms in the Taylor approximation. However, estimations of higher-order derivative terms will become more inaccurate given the successive estimations needed for each higher-order derivative. Noisy data would compound the difficulty in estimating accurate derivatives. Moreover, some functions will have a Taylor series that only converges within a specific range. Hence, a Taylor approximation's appropriateness will depend on the function being approximated, the point, $a$, used for the expansion and the distance between $a$ and the predicted point.

% Omer 16-05 - I think we need a stronger + clearer opening to this paragraph 
% Omer 16-05 - What do you mean by order of truncation ? is it mentioned here first time? 
% Bold X and Y are not defined? 
% maybe re-write line 142 to be more concise
The \textbf{LocalTaylor wrapper} models the mean prediction of $Y_T^i$ as the sum of i) explicit Taylor expansion terms and ii) a neural network adjustment function that uses Taylor expansion terms as features. This approach confronts the challenge of working out when Taylor approximations are useful (dependent on the noisiness of data, number of terms used in the Taylor approximation, and the gap over which predictions are made) by using a neural network to \textit{learn} how to use this information when making predictions. Giving features based on the Taylor expansion to the neural network can remove the hard-coded Taylor expansion terms if deemed not useful. Concretely,
\begin{equation}
     \mu_{Y_T^i} =\operatorname{LocalTaylor}(\operatorname{NeuralNet}, \mathbf{X}_C,X_T^i,\mathbf{Y}_C;p,q)
\end{equation}
where $p$ and $q$ are hyperparameters referring to the truncation order of the Taylor terms used for the hard-coded and neural network features, respectively. This is illustrated below for $p=1$ and $q=1$,
\begin{multline}
     \operatorname{LocalTaylor}(\operatorname{NeuralNet}, \mathbf{X}_C,X_T^i,\mathbf{Y}_C;1,1) = \\ Y_T^{n(i)} + \Delta X_T^{n(i)} D_T^{{n(i)}}
    + \\ \operatorname{Linear}\big(\operatorname{NeuralNet}(\mathbf{W},\mathbf{X}_C,X_T^i,\mathbf{Y}_C)\big)
\end{multline}
the first two terms are a hard-coded first-order Taylor approximation, and the third is the neural network adjustment function taking in Taylor expansion terms ($\mathbf{W}$) as features. Below, we describe the terms $Y_T^{n(i)},\Delta X_T^{n(i)},D_T^{n(i)}, \mathbf{W}$, and how they are created.

%Omer 16-05 - superscript "nn" is unclear
We introduce the RepresentationExtractor to create the terms required to compute the terms for a Taylor series approximation. It takes the hyperparameter $q$, and we show it for $q=1$:
\begin{multline}
     \operatorname{RepresentationExtractor}(X_T^i, \mathbf{X}_C, \mathbf{Y}_C;1) = \\ X_T^{n(i)}, \mathbf{X}_C^{n(I)},  Y_T^{n(i)}, \mathbf{Y}_C^{n(I)}, \\ \Delta X_T^{n(i)}, \Delta \mathbf{X}_C^{n(I)}, \Delta \mathbf{Y}_C^{n(I)}, \mathbf{D}_C, \mathbf{D}_C^{n(I)},D_T^{n(i)}
     \label{eq:rep_extractor}
\end{multline}
where $X^{n(i)}$  (the $i^{th}$ element of $\mathbf{X}^{n(I)}$) is the nearest already-seen neighbour of $X^\text{i}$, where the neighbour may come from either context set or previously seen values of the target set. Mathematically, $n(i) = \argmin_{i'\neq i, i'\in C \text{ or }i'<i} ||X^i - X^{i'}||$. The remaining terms are defined as $\Delta \mathbf{X}^{n(I)} = \mathbf{X} - \mathbf{X}^{n(I)}$, $\Delta \mathbf{Y}^{n(I)} = \mathbf{Y} -\mathbf{Y}^{n(I)}$ and $\mathbf{D} = \frac{\Delta \mathbf{Y}^{n(I)}}{\Delta \mathbf{X}^{n(I)}}$, a data-based approximation of the derivative at $\mathbf{X}$, and $\mathbf{D}^{n(I)}$ is an analogous data-based approximation of the derivative at $\mathbf{X}^{n(I)}$. These definitions hold for $\{\mathbf{X},\mathbf{Y}\}$ whether they are from the context or target sets. We handle ties in the $\argmin$ by randomly choosing one of the tied neighbours. We also positionally encode the $\mathbf{X}$ variables, similarly to \citet{Vaswani2017}. We then define $\mathbf{W}$ as the concatenation of these features,  $\mathbf{W} = \operatorname{Concat}[\text{i for i in} \operatorname{RepresentationExtractor}(X_T^i, \mathbf{X}_C, \mathbf{Y}_C;1)]$.

Inspiration for dealing with noisy data also comes from local smoothing techniques, such as LOESS, which deal with noise by adjusting the size of the data subsets used to fit their local functions. Similarly, we can smooth out noise using averages based on different data subsets. For example, instead of estimating derivatives based only on the nearest-neighbor point, we can take an average of derivatives where each is calculated based on different points near the estimation point. 

The estimation of derivatives  
% Although we have shown the approach for 1D functions, it 
can be generalised to higher-dimensions by using partial derivatives; we use it for our 2D regression experiments on the CelebA \cite{liu2015faceattributes} and EMNIST \cite{cohen2017emnist} datasets.

\subsection{MHA-X block}
\label{ref:method_mhax}

% We first motivate why it is useful to draw inspiration from looking at Gaussian Processes, then detail how GPs make predictions and finish by introducing the MHA-X block --- a block which allows predictions to be made in a similar manner as done for GPs.

It is worth drawing inspiration from how GPs model processes given they are well-liked probabilistic
tools used to model various continuous problems. In fact, they are even included in scikit-learn \cite{scikit-learn}. Although they model a restricted class of functions, based on their wide usage, it is evident that the classes of functions which they do model are important to users. 

For a GP, the mean of its predictions of $Y_T^i$ is simply a linear combination of the contextual information $Y_C$. This is seen from their predictive distribution for $Y_T^i$ at $X_T^i$ conditioned on $\mathbf{Y}_C$ and $\mathbf{X}_C$, $
Y_T^i \sim N(A,B)$, where 
\begin{equation}
   A = K(X_T^i,\mathbf{X}_C)K(\mathbf{X}_C,\mathbf{X}_C') \mathbf{Y}_C
   \label{eq:gp_a}
\end{equation}
where $K$ is a kernel specifying the covariance and is a function only of $\mathbf{X}$. Therefore, equation \eqref{eq:gp_a} is a linear weighting of $\mathbf{Y}_C$, with the weighting done by the kernel, which is a non-linear function of the $\mathbf{X}$ values.
% Omer 16-05 - need to explain what are values V
% Omer 16-05 - equation 14 and 15 are confusing due to X_c notation

GPs weigh contextual information in a contrasting manner to that in a standard GPT-style attention model, where non-linear combinations of $\{\mathbf{X}_C,\mathbf{Y}_C\}$ (referred to as values, $V$) are weighted based on non-linear functions of \textit{both} $\mathbf{X}_C$ and $\mathbf{Y}_C$ (referred to as the queries $Q$ and keys $K$) as follows
\begin{equation}
    \operatorname{Attention}(Q,K,V) = \operatorname{softmax}(\frac{QK^T}{\sqrt{d_k}}) V
    \label{eq:attention}
\end{equation}
where $V\in \mathbb{R}^{n_C \times d_v}$, $K \in \mathbb{R}^{n_C \times d_k}$, $Q \in \mathbb{R}^{n_C \times d_k}$. 

% Concretely,
% in standard attention we have values $V$, queries $Q$ and keys $K$ all three are functions of \textit{both} $\{\mathbf{X}_C,\mathbf{Y}_C\}$   language used in standard attention 

% Concretely, in NLP attention  values ($V$) get weighted according to scaled-dot product of queries ($Q$) and keys ($K$), which are typically non-linear functions of both $\mathbf{X}_C$ and $\mathbf{Y}_C$, as follows
% \begin{equation}
%     \operatorname{Attention}(Q,K,V) = \operatorname{softmax}(\frac{QK^T}{\sqrt{d_k}}) V
%     \label{eq:attention}
% \end{equation}
% where $V\in \mathbb{R}^{n_C \times d_v}$, $K \in \mathbb{R}^{n_C \times d_k}$, $Q \in \mathbb{R}^{n_C \times d_k}$. 

This standard attention's (MHA-XY) approach to weighting is important for language tasks as considerable weight should still be put on distant words if they have similar semantic meaning to closer-by words. But it is not an obvious requirement for regression tasks seen by how GPs are so useful despite this mechanism being absent.

% The standard Transformer's MHA-XY unit is not an obvious requirement for regression tasks. It may make it more challenging to learn models where a GP-style local smoothing would be more appropriate. For example, if emulating an ODE, it is closer-in-time values, which we suppose would be more beneficial for prediction. Just because a value in the context set is similar to the predicted points, if it is a large distance away, it may not be relevant. The MHA-XY unit would need to learn to ignore such relationships. 

We propose the \textbf{MHA-X block}, which allows the mean prediction of $Y_T^i$ to be modelled as a linear weighting of $\mathbf{Y}_C$, with the weighting being a non-linear function, learnt using a neural network, of just $\mathbf{X}_C$ and $X_T^i$.  

An MHA-X block with $N$ layers is composed of $N-1$, multi-head attention units
\begin{multline}
    \operatorname{MultiHead}(f(\mathbf{X}_C),f(\mathbf{X}_C),f(\mathbf{X}_C)) =  \\ \operatorname{Concat}(\text{head}_1,...,\text{head}_h)R^O 
\end{multline}
with $\text{head}_i = \operatorname{Attention}(QR_i^Q, KR_i^K,VR_i^V)$, where $f(\mathbf{X}_C)$ are non-linear functions only of features derived from $\mathbf{X}_C$, and $R_i^Q,R_i^K \in \mathbb{R}^{d_\text{model}\times d_k}$, $R_i^V\in \mathbb{R}^{d_\text{model}\times d_v}$ and $R^O\in \mathbb{R}^{hd_v\times d_\text{model}}$. The last layer of the MHA-X block is composed of the multi-head attention unit: \begin{equation}
    \text{MultiHead}(g(\mathbf{X}_C),g(\mathbf{X}_C),\mathbf{Y}_C)
\end{equation}
where $g(\mathbf{X}_C)$ is the output of the previous layer in the MHA-X block.

\subsection{Training}
\label{method:training}
% Omer 16-05 - need better sign-posting
% Omer 16-05 - if we don't care about efficency why mention?
% Omer 16-05 - I think we need to re-write lines 198 to 202 it's too full of notation 
% Omer 16-05 - Is line 209-210 correct?? 
Our model is trained by maximising the likelihood, $\mathrm{Pr}(\mathbf{Y}_T|\mathbf{X}_T,\mathbf{X}_C,\mathbf{Y}_C)$. Two key differences exist between our training and a standard transformer-decoder model like GPT. The first is using a shuffling method to encourage consistency through training. For tasks where we wish to prioritise target-equivariance, we use a shuffling and Expectation approach similar to \citet{pmlr-v162-nguyen22b}: during training, we maximise the likelihood of $p(\mathbf{Y}_{\pi(T)}|\mathbf{X}_{\pi(T)},\mathbf{Y}_C,\mathbf{X}_C)$ where $\pi$ are permutations of the target set which are chosen randomly for each training batch. The Taylorformer enforces context invariance (equation \eqref{eq:context in}) through its architecture.

The second difference is an approach to prevent data leakage when training to predict in arbitrary orders. To predict points in arbitrary orders, we need a mechanism to query points at arbitrary $\mathbf{X}$ locations without revealing the corresponding $\mathbf{Y}$ value during training. The classic GPT-style mask is not capable of this. \citet{pmlr-v162-nguyen22b} introduced a mask and target-padding approach that deals with this. Our process is an alternative approach. The main difference is that the number of multiplications required for the scaled dot-product in equation \eqref{eq:attention} is $O((n_C + 2n_T)^3)$ for \citet{pmlr-v162-nguyen22b}, whereas ours is only $O((n_C+n_T)^3)$ operations, though we reiterate that efficiency is not our work's focus. Our mechanism combines separate keys and queries for the attention mechanism with specific masking. The process is different for MHA-X and MHA-XY and is detailed below.

\paragraph{MHA-XY}

For the first layer in the MHA-XY block, there are $n_C + n_T$ queries, keys and values. The queries are 
\begin{equation}
Q_i =
\begin{cases}
(X^{\text{fe},i},Y^{\text{fe},i},Y^{\text{seen},i},1) & \text{if }i\in C\\
(X^{\text{fe},i},0,Y^{\text{seen},i},0) & \text{if }i\in T
\end{cases}
\label{eq:query_xy}
\end{equation}
where $X^{\text{fe},i}$ are features which are only functions of $X^i$ and $X^j$ for $j \in C \cup \{j:j<i\}$, $Y^{\text{fe},i}$ are features which are functions of $Y^i$, $X^i$, $Y^j$ and $X^j$ for $j \in C \cup \{j:j<i\}$, and $Y^{\text{seen},i}$ are features which are functions of $X^i$, $Y^j$ and $X^j$ for $j \in C \cup \{j:j<i\}$. For this work, $X^{\text{fe},i} = [X^{i},X^{n(i)},\Delta X^{n(i)}]$, $Y^{\text{fe},i} = [Y^i,\Delta Y^{n(i)},D^{i}]$ and $Y^{\text{seen},i} = [Y^{n(i)},D^{n(i)}]$, where these terms are defined in section \ref{method:taylor}. We mask $Y^{\text{fe},i}$ so that when making predictions for the target set, we can query by $Q_i$ (as we need to know the $X$ information of the point we are looking to predict) without revealing the target value during training. The final item is a label indicating whether the variables $Y^{\text{fe},i}$ which contain $Y^i$ are masked (set to zero) or not. The keys and values are
\begin{equation}
    K_i = V_i = (X^{\text{fe},i},Y^{\text{fe},i},Y^{\text{seen},i},1)
    \label{eq:value_xy}
\end{equation}

The masking mechanism is designed so that: (1) the context points only attend to themselves and (2) target points only attend to the previous target points and the context points. Figure \ref{fig:architecture}b shows an example mask for $n_C = 3 $ and $n_T = 3$.

% \begin{figure}[ht]
% \includegraphics[width=5cm]{figures/masking.pdf}
% \caption{An example mask for $n_C = 3$ and $n_T =3$. The context points ($Q_i$ for $i \in C$) attend to themselves ($K_i$ for $i \in C$). The masked target points ($Q_i$ for $i \in T$) attend to context points ($K_i$ for $i \in C$) and previous target points ($K_i$ for $i' < i$). When a point is in the target set, we can see all context points and target points we've encountered already. For example, to predict $Y^i=6$, we use the last row from the masking matrix. All three context points and previous observations are not masked.}       
% \label{maskX}
% \end{figure}

% \begin{wrapfigure}{l}{0.5\textwidth}
%   \begin{center}
%     \includegraphics[width=0.8\textwidth,trim={0cm 5cm 4cm 4.2cm},clip]{figures/atp_masking_figure.pdf}
%   \end{center}

% \end{wrapfigure}

For subsequent layers in MHA-XY, the queries, keys and values are the outputs of the previous MHA-XY layers.

\paragraph{MHA-X}

For the first N-1 layers, the inputs are the outputs of the previous layer, where the inputs to the first layer are
\begin{equation}
    Q_i = K_i = V_i = X^{\text{fe},i}
\label{eq:query_x}
\end{equation}
% where $X^{fe,i}$ is specified above.
For the final layer, the inputs are $Q_i = K_i = O_i$, where $O_i$ is the output of the previous MHA-X layer, and $V_i = Y_i$. The same mask as for MHA-XY is used for MHA-X.

\section{Experiments}
% **** DONT FORGET GRU RESULTS ****
% We support our claim for a better (in terms of results) model for processes concerning the NP literature by testing on their key tasks ---  1D regression and 2D regression (tested on image completion) --- and concerning forecasting literature by testing on three key tasks of time series --- electricity consumption (measured in KWH), the load of electricity transformers and exchange rate. 

To assess the Taylorformer as a model for continuous processes, we evaluate on key tasks in the NP literature: 1D regression and 2D regression (tested on image completion). To assess it on time-series, we evaluate it on three key forecasting tasks: electricity consumption, the load of electricity transformers and exchange rate. 

For these experiments, it was empirically better to truncate the explicit Taylor expansion terms to the zeroth order (setting $p=0$) and keep the first order terms in $\mathbf{W}$ (setting $q=1$) with using the approximate derivative of the nearest neighbour. Therefore, we model the mean of $Y_T^i$ as $
     {\mu_{Y_T^i} =\operatorname{LocalTaylor}(\operatorname{NeuralNet}, \mathbf{X}_C,X_T^i,\mathbf{Y}_C;0,1)}$. 
     We approximate the expectation using a single-sample Monte Carlo.

The Taylorformer was trained on a 32GB NVIDIA V100S GPU. Further implementation details are below.

\subsection{Neural Process tasks}

\paragraph{Datasets} 
% We use datasets corresponding to 1D or 2D regression tasks. 
% For 1D regression, we generate three datasets corresponding to a meta-learning setup --- each dataset contains 100K sequences with $\mathbf{X} \sim U[-2,2]$  ($\mathbf{X} \in \mathbf{R}^{200}$). We query Gaussian Process (GPs) at those $\mathbf{X}$ to give $\mathbf{Y}$ with a random selection of kernel hyper-parameters as specified in \ref{appendix_experimental_details}. Each dataset uses one of  RBF, Mat\'ern, and a Periodic kernel for the GP generation. Hold-out dataset includes sequences generated in the same manner as for training data. 
For 1D regression, we generate three datasets corresponding to a meta-learning setup --- each dataset contains 100K sequences with $\mathbf{X} \sim U[-2,2]$. We query Gaussian Process (GPs) at those $\mathbf{X}$ to give $\mathbf{Y}$ with a random selection of kernel hyper-parameters as specified in Appendix \ref{appendix_experimental_details}. Each dataset uses either the RBF, Mat\'ern or Periodic kernel for GP generation. The hold-out dataset includes sequences generated in the same manner as the training data.

% For 2D regression, we use three datasets. The first two come from the balanced EMNIST \cite{cohen2017emnist} dataset, which contains monochrome images of 47 handwritten numbers and letters. Each pixel $y_i \in \mathbf{R}^1$ is scaled to be in [-0.5, 0.5] and $x_i \in \mathbf{R}^2$ with each dimension scaled to be in [-1, 1]. The first dataset EMNIST-seen includes in the hold-out set unseen images from the same classes seen during training. The second dataset EMNIST-unseen includes images from unseen (during training) classes (10-46, corresponding to handwritten letters) in the hold-out set during training. The third dataset is CelebA \cite{liu2015faceattributes}: coloured images of celebrity faces with each pixel $y_i \in \mathbf{R}^3$ with each dimension scaled to be in [-0.5, 0.5] and $x_i \in \mathbf{R}^2$ with each dimension scaled to be in [-1, 1].

For 2D regression, we use two datasets. The first comes from the balanced EMNIST \cite{cohen2017emnist} dataset, which contains monochrome images. Each pixel $y_i \in \mathbf{R}^1$ is scaled to be in [-0.5, 0.5] and $x_i \in \mathbf{R}^2$ with each dimension scaled to be in [-1, 1]. We train on only a subset of available classes (0-9, corresponding to handwritten numbers). There are two hold-out sets, the first includes unseen images from the same classes seen during training. The second includes images from unseen classes (10-46, corresponding to handwritten letters). The second dataset is CelebA \cite{liu2015faceattributes}: coloured images of celebrity faces with each pixel $y_i \in \mathbf{R}^3$ with each dimension scaled to be in [-0.5, 0.5] and $x_i \in \mathbf{R}^2$ with each dimension scaled to be in [-1, 1].
% \vspace{-0.8em}
\paragraph{Implementation details}  
% We train all datasets to maximise average log-likelihood --- a metric combining predictive accuracy with uncertainty --- of the target data, $\frac{1}{n_T} \log p(\mathbf{Y}_T|\mathbf{X}_T,\mathbf{X}_c,\mathbf{Y}_c)$. Models were
% trained for 250K, 71K and 125K iterations with 32, 64, and 64 batch sizes for 1D regression, EMNIST and CelebA, respectively. Optimization was done using Adam with a $10^{-4}$ learning rate. For 1D-regression $n_C \sim U[3,97]$ and $n_T = 100 - n_C$. For EMNIST and CelebA  $n_C \sim U(6,200)$ and $n_T \sim U(3,197)$. Full implementation details are found in the appendix.
Models were trained for 250K, 71K and 125K iterations with 32, 64, and 64 batch sizes for 1D regression, EMNIST and CelebA, respectively. Optimization was done using Adam with a $10^{-4}$ learning rate. For 1D-regression, $n_C \sim U[3,97]$ and $n_T = 100 - n_C$. For EMNIST and CelebA  $n_C \sim U(6,200)$ and $n_T \sim U(3,197)$. Full implementation details are found in the Appendix.

% \paragraph{Baselines}
% The meta-learning 1D-regression and image completion experiments were used extensively in the NP literature \cite{Garnelo2018NeuralP,gordon2019convolutional,Kim2019AttentiveNP,lee2020bootstrapping,pmlr-v162-nguyen22b,wang2020doubly}, and we compare to NP \cite{Garnelo2018NeuralP}, ANP \cite{Kim2019AttentiveNP} and TNP \cite{pmlr-v162-nguyen22b}. 

% \paragraph{Results}
% Table \ref{np_table} shows that ATP improves on the other methods for 5 out of 6 tasks. Figure \ref{figure: loss_1d_reg} shows that for all  1D regression tasks, the validation loss is lowest for the ATP model. For the 2D regression, we perform best on both EMNIST tasks, but there is no improvement over the TNP for CelebA. It is unclear what properties of the CelebA dataset prevent performance improvement. We hypothesise that noise in the dataset may mean our local smoothing approach is less practical when a single sample is used and may point to the need to take further samples.

\paragraph{Results.}
The meta-learning 1D-regression and image completion experiments were used extensively in the NP literature \cite{Garnelo2018NeuralP,gordon2019convolutional,Kim2019AttentiveNP,lee2020bootstrapping,pmlr-v162-nguyen22b,wang2020doubly}, and we compare to NP \cite{Garnelo2018NeuralP}, ANP \cite{Kim2019AttentiveNP} and TNP \cite{pmlr-v162-nguyen22b}. Table \ref{np_table} shows that Taylorformer improves on the other methods for 5 out of 6 tasks. Figure \ref{figure: loss_1d_reg} in Appendix \ref{appendix:further_results} shows that for all  1D regression tasks, the validation loss is lowest for the Taylorformer. For the 2D regression, we perform best on both EMNIST tasks, but there is no improvement over the TNP for CelebA. We hypothesise that noise in the CelebA dataset may benefit from taking averages of derivatives.

\begin{table*}[tp]%
\caption{Log-likelihood on a test set. Higher is better. Taylorformer outperforms the SOTA TNP in 5/6 1D and 2D regression tasks. Each model is run five times to report log-likelihood with standard deviation results.}
\centering% 
\small
\begin{tabular}{llcccc}
\toprule%
& &&Log-Likelihood\\
\multicolumn{2}{l}{}  & \textbf{Taylorformer}   &  TNP & ANP & NP \\
% \multicolumn{2}{l}{Regression task} &&&&\\
% \cmidrule(lr){3-3} \cmidrule(l){4-4} \cmidrule(l){5-5} \cmidrule(l){6-6}
\otoprule%
\multicolumn{1}{|l|}{\multirow{5}{*}{\rotatebox[origin=r]{90}{1D}}} & 
                            RBF  & {\boldmath$4.13 \pm  0.03$}& $3.50 \pm 0.05$ & $0.96 \pm 0.02$
                            
                           &  $0.24 \pm 0.01$\\ 
                          \multicolumn{1}{|l|}{}     \\
                   \multicolumn{1}{|l|}{}      & Matern & {\boldmath$3.87 \pm 0.05$}  & $3.22 \pm 0.05$ & $1.07 \pm 0.01$& $0.40 \pm 0.02$ \\ \multicolumn{1}{|l|}{} 
                               
                              \\ \multicolumn{1}{|l|}{} & Periodic & {\boldmath$0.31 \pm 0.02$} & $0.13 \pm 0.04$ & $ -0.85 \pm 0.05$& $-1.56 \pm 0.00$
                              \\\midrule
\multicolumn{1}{|l|}{\multirow{5}{*}{\rotatebox[origin=r]{90}{2D}}}& 

                            EMNIST-seen  &  {\boldmath$2.18 \pm 0.01$}
                            & $1.59 \pm 0.00$ & $0.82 \pm 0.04$& $0.62 \pm 0.01$   \\ \multicolumn{1}{|l|}{} 
                                \\ \multicolumn{1}{|l|}{} 
                              & EMNIST-unseen  &{\boldmath$1.90 \pm 0.03$}& $1.47 \pm 0.01$ & $0.68 \pm 0.08 $ & $0.44 \pm 0.00$\\ \multicolumn{1}{|l|}{} 
                              \\ \multicolumn{1}{|l|}{}  & CelebA &  $4.00 \pm 0.02$& {\boldmath$ 4.13 \pm 0.02$}& $2.78 \pm 0.03$ & $2.30 \pm 0.01$\\

\bottomrule 
\end{tabular}
\label{np_table}
\end{table*}

\subsection{Forecasting}

\paragraph{Datasets} 
% \begin{noindlist}
%   \item [1]. Hourly electricity \footnote{\url{https://drive.google.com/file/d/1jinfTAApPyuyvW1P1hUDpI3rl0Jq8in1/view?usp=share_link}} consumption data from 2013 to 2014. We pick client number 322 out of the 370 available clients. 
%   \item[2]. 15-minute Electricity transformers load (ETT) \cite{Zhou2020InformerBE}.  
%   \item [3].Daily exchange rate \cite{Lai2017ModelingLA} from 1990-2016. We pick one out of the eight countries in the data. 

% \end{noindlist}
% We use three datasets: 1) Hourly electricity (KWH) \footnote{\url{https://drive.google.com/file/d/1jinfTAApPyuyvW1P1hUDpI3rl0Jq8in1/view?usp=share_link}} consumption data from 2013 to 2014. We pick client number 322 out of the 370 available clients. 2) 15-minute Electricity transformers load (ETT)  \cite{Zhou2020InformerBE} and 3) Daily exchange rate \cite{Lai2017ModelingLA} from 1990-2016. We pick one out of the eight countries in the data. 

We use three datasets: 1) hourly electricity consumption data from 2013 to 2014.\footnote{\url{https://drive.google.com/file/d/1jinfTAApPyuyvW1P1hUDpI3rl0Jq8in1/view?usp=share_link}} We pick client number 322 out of the 370 available clients. 2) 15-minute Electricity transformers load (ETT)  \cite{Zhou2020InformerBE} and 3) Daily exchange rate \cite{Lai2017ModelingLA} from 1990-2016 (we pick the country denoted $OT$). For each dataset, we run four experiments in which each sequence has a fixed length context set, $n_C = 96$, (with randomly selected starting position) and a prediction length $n_T \in \{96, 192, 336, 720\}$. We standardise $y_i \in \mathbf{R}$ for all experiments. $x_i \in \mathbf{R}$ is scaled to be in [-1, 1] in each sequence --- only relative time matters to us.

% For each of the three datasets, we run four experiments in which each sequence has a fixed $n_C = 96$ length context set (with randomly selected starting position) and a prediction length $n_T \in \{96, 192, 336, 720\}$. We have standardised $y_i \in \mathbf{R}$ for all experiments. $x_i \in \mathbf{R}$ is scaled to be in [-1, 1] in each sequence --- only the relative time-points matter for us. 

\paragraph{Implementation details}  
% We train all datasets to maximise average log-likelihood. Models were
% trained for 40K iterations with batch sizes of 32. Optimization was done using Adam with a $10^{-4}$ learning rate. All datasets are split chronologically into training, validation and test sets by the ratios 72:8:20, 69:11:20, and 69:11:20, respectively. Full implementation details are in Appendix A.
Models were trained for 40K iterations with batch sizes of 32. Optimization was done using Adam with a $10^{-4}$ learning rate. All datasets are split chronologically into training, validation and test sets by the ratios 72:8:20, 69:11:20, and 69:11:20, respectively. Full implementation details are in Appendix A.

% \paragraph{Baselines}
% There are a number of forecasting models not based on attention. Amongst classical methods, ARIMA is popular. We do not include this in our evaluation as Autoformer \cite{Wu2021AutoformerDT} and Informer \cite{Zhou2020InformerBE} are both shown to beat it in forecasting. RNN-based models, including LSTMs \cite{hochreiter1997long} and GRUs \cite{cho2014learning}, have also been used for forecasting [****cite****] . We compare to a GRU in our results.

% The electricity, ETT and exchange rate were used extensively in the forecasting literature, and we compare to Autoformer \cite{Wu2021AutoformerDT}, Informer \cite{Zhou2020InformerBE}, TNP \cite{pmlr-v162-nguyen22b} and a GRU \cite{cho2014learning} model. For Autoformer and Informer, we keep the exact setup used in the papers concerning network and training parameters so the model is trained on mean-squared-error. We train TNP and GRU to maximise log-likelihood, and we set the number of network parameters to match ours since their paper does not implement these tasks. Further, we run a hyper-parameter search for TNP and GRU. We do not compare to the ARIMA model as the Autoformer and Informer have already performed better on these datasets.   

\paragraph{Results}
The electricity, ETT and exchange rate are used extensively in the literature, and we compare to Autoformer \cite{Wu2021AutoformerDT}, Informer \cite{Zhou2020InformerBE} and TNP \cite{pmlr-v162-nguyen22b}. For Autoformer and Informer, we keep the exact setup used in the papers concerning network and training parameters so the model is trained on mean-squared-error. We train TNP to maximise log-likelihood, and we set the number of network parameters to match ours since their paper does not implement these tasks. Further, we run a hyper-parameter search for TNP. We do not compare to the classic ARIMA model as it is out-performed by Autoformer and Informer on these datasets.   

Table \ref{table:forecasting_table} shows that Taylorformer is notably better in terms of MSE for all forecasting tasks. We outperform the autoregressive TNP and the batch-generating Autoformer and Informer with 21-34\%, 95-99\% and 14-18\% reductions in MSE compared to the closest model for the ETT, exchange and electricity tasks, respectively. Examples of our generated sequences can be seen in Figures \ref{fig:samples_ett} and \ref{figure:electricity_samples} (Appendix). The likelihood results show a similar trend and are provided in Appendix \ref{appendix:further_results}.

% \paragraph{Results}
% Table \ref{table:forecasting_table} shows that ATP noticeably improves in MSE on the other methods for all forecasting tasks. We outperform the autoregressive TNP and the batch-generating Autoformer and Informer with 8-12\%, 75-86\% and 17-23\% reductions in MSE compared to the closest model for the ETT, exchange and electricity tasks, respectively. The quality of our generated sequences is high and can be seen in \ref{figure:electricity_samples} compared to the TNP-generated sequences.

% \textbf{Training}. For EMNIST, only images from classes 0-9 are seen in training. For both datasets, $n_{C}$ and $n_{T}$ are chosen randomly from the same distributions used in the evaluation.
\begin{table*}[tp]%
\caption{MSE on a test set. Lower is better. Taylorformer outperforms three state-of-the-art models on three forecasting tasks (details in text). Each task is trained and evaluated with different prediction lengths $n_{T} \in \{96, 192, 336, 720\}$ given a fixed 96-context length. Each model is run five times to report log-likelihood with standard deviation results.}
\centering% 
\small
\begin{tabular}{llcccc}
\toprule%
% \multicolumn{2}{l}{Models}  
&&&MSE\\
& \multicolumn{2}{c}{\textbf{Taylorformer}}   &  \multicolumn{1}{c}{TNP} & \multicolumn{1}{c}{Autoformer} & \multicolumn{1}{c}{Informer}\\ 
\otoprule%
\multicolumn{1}{|l|}{\multirow{4}{*}{\rotatebox[origin=r]{90}{ETT}}} & 
                            96  &  \boldmath{$0.00030 \pm 0.00001$} &$0.00041 \pm 0.00004$ & $0.43 \pm 0.21$&
                              $0.40 \pm 0.20$
                              \\ \multicolumn{1}{|l|}{} 
                              & 192 & \boldmath{$0.00029 \pm 0.00000$} & $0.00044 \pm 0.00008$&$0.57 \pm 0.26$
                              &$0.65 \pm 0.30$\\ \multicolumn{1}{|l|}{} 
                              & 336 & \boldmath{$0.00030 \pm 0.00001$} & $0.00038 \pm 0.00001$ & $0.67 \pm 0.32$&
                              $0.72 \pm 0.34$
                               \\ \multicolumn{1}{|l|}{} 
                              & 720 & \boldmath{$0.00029 \pm 0.00000$} &$0.00039 \pm 0.00004$ & $0.87 \pm 0.42$& $ 0.85 \pm 0.40$\\\midrule
\multicolumn{1}{|l|}{\multirow{4}{*}{\rotatebox[origin=r]{90}{Exchange}}}& 96  & \boldmath{$0.002 \pm 0.000$} & $0.040 \pm 0.030$& $0.41 \pm 0.20$& 
$0.57 \pm 0.26 $ \\
                            \multicolumn{1}{|l|}{}   & 192 & \boldmath{$0.002 \pm 0.000$} & $0.079 \pm 0.035$ & $0.59 \pm 0.28$& $1.53 \pm 0.80$\\
                             \multicolumn{1}{|l|}{}  & 336 & \boldmath{$0.002 \pm 0.000$} & $0.067 \pm 0.025$ &$0.96 \pm 0.46$& $3.28 \pm 1.4$\\
                           \multicolumn{1}{|l|}{}    & 720 & \boldmath{$0.001 \pm 0.000$}  & $0.152 \pm 0.091$ &$1.08 \pm 0.51$& $1.47 \pm 0.76$\\[1ex]\midrule 
\multicolumn{1}{|l|}{\multirow{4}{*}{\rotatebox[origin=r]{90}{Electricity}}}

                              & 96    
                              & \boldmath{$0.036 \pm 0.000$}  
                              
                              &
                              $0.042 \pm 0.002$
                              &
                              
                              $0.130 \pm 0.039$
                              &
                              
                              $0.133 \pm 0.005$                               
                              \\  \multicolumn{1}{|l|}{} 
                              & 192 & 
                             \boldmath{$0.037 \pm 0.000$}
                
                              &
                              $0.045 \pm 0.001$
                 
                              & 
                              $0.117 \pm 0.001$
             
                              &
                              $0.148 \pm 0.011$ 
                               
                              \\ \multicolumn{1}{|l|}{} 

                              & 336 &  
                              \boldmath{$0.040 \pm 0.001$}
                              & $0.048 \pm 0.002$ 
                              & $0.143 \pm 0.011$
                              & $0.150 \pm 0.013$
                              \\ \multicolumn{1}{|l|}{} 

                              & 720 & 
                              \boldmath{$0.039 \pm 0.001$}  
                              &$0.048 \pm 0.003$
                              &$0.216 \pm 0.040$
                              &$0.129 \pm 0.012$ 
                              \\[2ex]

\bottomrule 
\end{tabular}

\label{table:forecasting_table}
\end{table*}

\subsection{Ablation study}
% We study what contributions are key to the final model performance by performing a 1D regression task on four equally sized models: a base transformer-decoder, a model with only a MHA-X, a model with only LocalTaylor wrapper and a model with both. Results presented in figure \ref{figure:ablation} in Appendix A indicate that the combination of both contributions yields the best outcome.
We study what contributions are key to the final model performance by performing a 1D regression task on four equally sized models: a base transformer-decoder (just MHA-XY units), a model with only MHA-X, a model with the LocalTaylor wrapped around the base model, and a model with both contributions. Results in Figure \ref{figure:ablation} indicate that the combination of both contributions yields the best outcome.

\begin{figure}[h]
\vskip -0.17in
\begin{center}
\centerline{\includegraphics[width=\columnwidth,trim={0 0 0 0cm},clip]{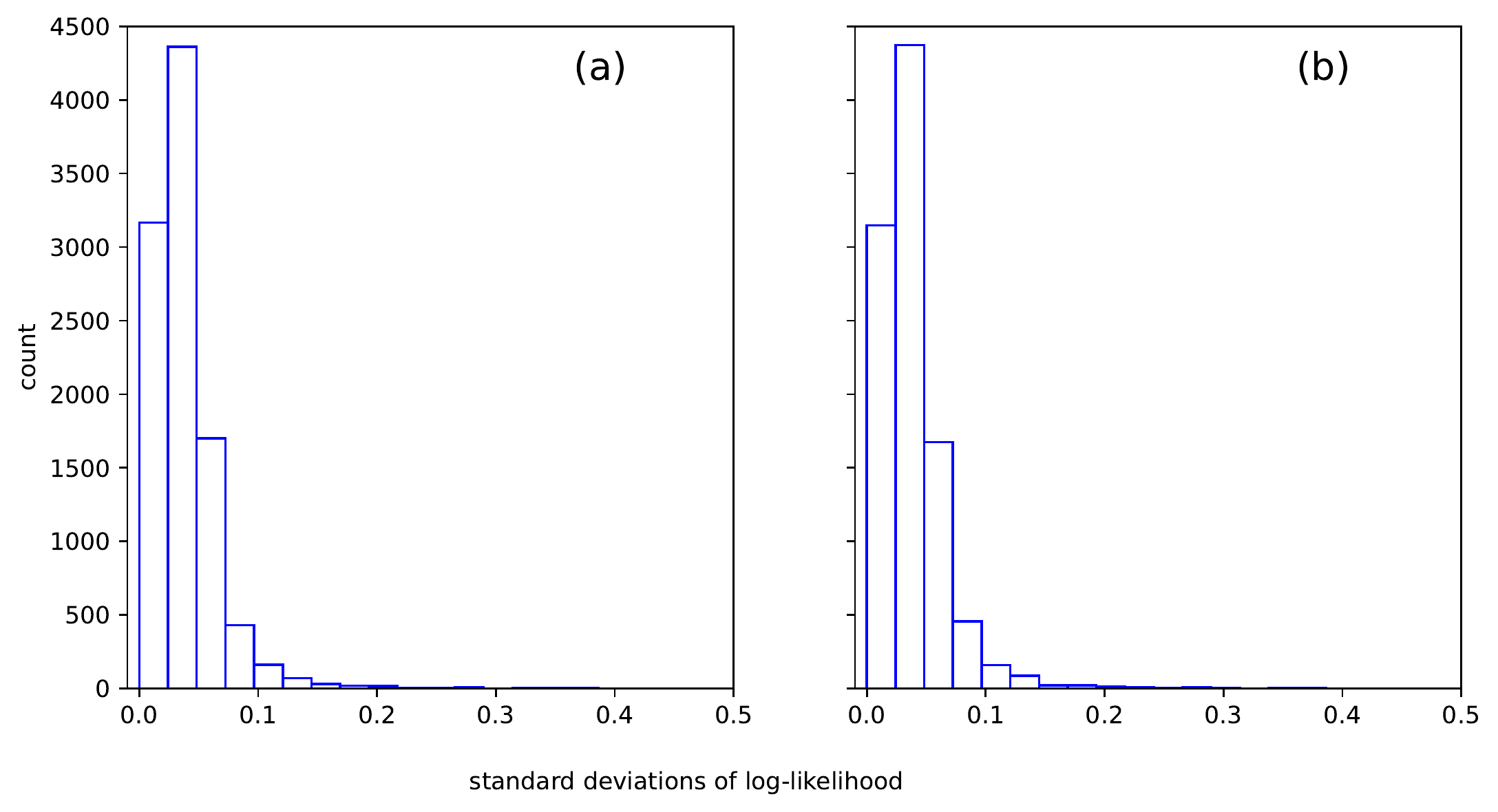}}
\caption{By shuffling the target variables given the context during training, we drive the model to be approximately target equivariant. If all log-likelihood scores are equal, their standard deviation will be zero. The histograms show these standard deviations for training with (a) one permuted sample and (b) five permuted samples on the RBF task. We can see that the models are `close' to consistency. Furthermore, for the specific RBF task, one permuted sample during training seems suitable to drive consistency.}    

\label{figure:consistency}
\end{center}
% \vskip -0.2in
\end{figure}
% \vspace{-0.9em}

\subsection{Evaluating consistency}
\label{section:evaluating_consistency}
Our model attempts to approximate target equivariance via shuffled training. We evaluate the target equivariance (equation \eqref{eq:target_equi}) of Taylorformer by computing the log-likelihood for sequences with permuted target points and taking the standard deviation. If consistent, all the likelihoods would be the same and the standard deviation would be zero. 

Concretely, for each of the H sequences in the hold-out set, the target set is permuted 40 times. The log-likelihoods of these permuted sequences are calculated, and the standard deviations of these are computed. This results in an array of H standard deviations. This can then be visualised with a histogram. A consistent model would show a histogram only composed of points at zero (as the calculated log-likelihoods of a given sequence would not change for the differing target set permutations).

We show the results of this procedure for the RBF experiment in Figure \ref{figure:consistency}a. Our model does not achieve consistency but comes close. The mean of the standard deviations is 0.038, with a 95\% confidence interval of [0.002, 0.105]. We compare this to an identical model trained with 5 permutation samples for each training sequence (instead of 1) to better approximate equation. Figure \ref{figure:consistency}b shows that there is no obvious improvement from this, which is supported by the mean of the standard deviations also being 0.038, with a 95\% confidence interval of [0.002, 0.110].

% Consistency is desirable, as discussed in Section \ref{section:shuffling}. Our model attempts to approximate consistency via shuffled training. Here is how we measure how well it achieves equivariance concerning the target set: for each of the $ B $ sequences in the hold-out set, the target set is permuted 40 times. The log-likelihoods of these permuted sequences are calculated, and the standard deviations of these are computed, resulting in an array of $B$ standard deviations. The array can then be visualised with a histogram. A consistent model would show a histogram composed of points at zero (as the calculated log-likelihoods of a given sequence would not change for the differing target set permutations). 

% We show the results of this procedure for the RBF experiment in Figure \ref{figure:consistency}a. Our model does not achieve consistency but comes close. The mean of the standard deviations is $0.038$, with a 95\% % confidence interval of $[0.002,0.105]$. We compare this to an identical model trained with five samples (instead of 1) to better approximate equation \eqref{eq::consistency} in training. Figure \ref{figure:consistency}b shows no noticeable improvement, which is supported by the mean of the standard deviations also being $0.038$, with a 95\% % confidence interval of $[0.002,0.110]$.

\section{Discussion}

\bibliography{example_paper}
\bibliographystyle{icml2023}

%%%%%%%%%%%%%%%%%%%%%%%%%%%%%%%%%%%%%%%%%%%%%%%%%%%%%%%%%%%%%%%%%%%%%%%%%%%%%%%
%%%%%%%%%%%%%%%%%%%%%%%%%%%%%%%%%%%%%%%%%%%%%%%%%%%%%%%%%%%%%%%%%%%%%%%%%%%%%%%
% APPENDIX
%%%%%%%%%%%%%%%%%%%%%%%%%%%%%%%%%%%%%%%%%%%%%%%%%%%%%%%%%%%%%%%%%%%%%%%%%%%%%%%
%%%%%%%%%%%%%%%%%%%%%%%%%%%%%%%%%%%%%%%%%%%%%%%%%%%%%%%%%%%%%%%%%%%%%%%%%%%%%%%
\newpage
\appendix
\onecolumn
\section{Experimental Details}
\label{appendix_experimental_details}

\begin{table}[tp]%
\caption{Network and training details for forecasting tasks ETT/ exchange/ electricity. If a cell contains one value then all tasks share the same parameter and otherwise the cell is split by a slash respectively. ${}^{*}$ indicates that early stopping was used. For the electricity dataset the total number of parameters for Autoformer and Informer is reduced (relative to original implementation) to improve their results and match the number of Taylorformer parameters.}

\label{Model_parameters}\centering% 
\small
\begin{tabular}{l|cccc}
\toprule%
Models  & \textbf{Taylorformer}  & TNP& Autoformer & Informer\\
\cmidrule(lr){1-1} \cmidrule(l){2-2} \cmidrule(l){3-3} \cmidrule(l){4-4} \cmidrule(l){5-5} 
Network parameters     \\\cmidrule(lr){1-1} 
Dropout rate  & 0.05/0.05/0.25 & 0.05/0.1/0.25 & 0.05 & 0.05                             \\\\
\multirow{3}{*}{\# Layers}     & attention heads & attention heads & 2 encoder,  &  2 encoder,         \\
& 6/8/6 & 7/6/4& 1 decoder & 1 decoder\\
\\
\# Parameters  & $90K$/$90K$/$200K$ & $90K$/$90K$/$200K$ & $42m$/$42m$/$200K$ & $45m$/$45m$/$200K$                            
                              \\\midrule
Training parameters     \\\cmidrule(lr){1-1} 
\# Iterations & $40K^*$ & $40K^*$ & $15K^*$ & $15K^*$ \\ 
Batch size    &  32 & 32& 32&32                            \\
Learning rate  &  $3e^{-4}$ & $3e^{-4}$ & $1e^{-4}$ & $1e^{-4}$                           \\
Optimizer & $\operatorname{Adam}$ & $\operatorname{Adam}$ & $\operatorname{Adam}$& $\operatorname{Adam}$                              
                              \\\midrule
Hardware        \\\cmidrule(lr){1-1}                       
Processor  & \multicolumn{2}{|c}{32GB NVIDIA V100S GPU}  &  \multicolumn{2}{|c}{120GB NVIDIA T4x4 GPU}                                                       \\[2ex]

\bottomrule 
\end{tabular}

\end{table}

The NP, ANP and TNP architectures follow closely from those used by \cite{pmlr-v162-nguyen22b}. The TNP hyperparameters are the same as in \cite{pmlr-v162-nguyen22b} for the 1D regression and image completion tasks. For the electricity forecasting task, we set the TNP dropout to 0.25 to reduce overfitting. Specifics for the ANP and NP are provided below. The TNP, NP and ANP all use Relu activation functions for their Dense layers. 

The Autoformer and Informer architectures follow exactly those from the official code base of \cite{Wu2021AutoformerDT} --- code can be found in \url{https://github.com/thuml/Autoformer}. The number of parameters in both models are reduced to make a fair comparison with our model. This practically means that we set d\_model $= 20$ and dff $= 512$ (notation as used in the original work). 

In all cases, during training we monitored performance on a validation set and saved the model weights for the iterations they performed best (in terms of validation log-likelihood). 

\subsection{1D Regression}

\paragraph{Data generation.} For a GP with a given kernel, first we sampled random kernel hyparameters (distributions are given below). Next, we generated $\mathbf{X} \sim U[-2,2]$, where $\mathbf{X} \in \mathbb{R}^{200\textrm{x}1}$ and then queried the GP at those locations to give $\mathbf{Y}$. From this, $n_C$ context pairs and $n_T$ target ones were randomly selected, with $n_C \sim U(3,97)$ and $n_T = 100 - n_C$. This process was repeated to generate all the sequences. The hyperparameters were drawn from the following distributions: for the RBF kernel, $k(x,x') = s^2 \exp(-|x-x'|^2 / 2l^2)$, $s \sim U(0.1,1.0)$ and $l \sim U(0.1,0.6)$. For the Mat\'ern 5/2, $k(x,x') = (1+\sqrt{5}d/l + 5d^2/(3l^2))\exp(-\sqrt{5}d/l), d = |x-x'|$, $l \sim U(0.3,1.0)$. For the periodic kernel, $k(x,x') = \exp(-2\sin^2(\pi |x-x'|^2 /p)/l^2)$, $l \sim U(0.1,0.6)$ and $p \sim U(0.5,1.0)$.

\paragraph{Training and testing.} Three training sets (for each kernel) were generated, comprising 100,000 sequences each. The test set (for each kernel) comprised 10,000 unseen sequences. All models were trained for 250,000 iterations, with each batch comprising 32 sequences. We used Adam with a learning rate of $10^{-4}$. 

\paragraph{Architectures.}  For the NP, the latent encoder has 3 dense layers before mean aggregation, and 2 after it. The deterministic encoder has 4 dense layers before mean aggregation. The decoder has 4 dense layers. All dense layers are of size 128. For the ANP, the latent encoder and decoder are the same as for the NP. The deterministic encoder has an additional 2 dense layers to transform the key and queries. The multihead attention has 8 heads. All dense layers are of size 128. 

\subsection{Image completion}

\paragraph{Data generation.}

For a given image, we sampled the 2D coordinates of a pixel, and rescaled to [-1,1] to compose $\mathbf{X}$ and rescaled the pixel value to [-0.5,0.5] to compose $\mathbf{Y}$. The number of context, $n_C$, and target, $n_T$, points are chosen randomly for each image, with  $n_C \sim U(6,200)$ and $n_T \sim U(3,197)$. 

\paragraph{Training and testing.} 

For EMNIST, all models were trained for 200 epochs (71,000 iterations with a batch size of 64). For CelebA, all were trained for 45 epochs (125,000 iterations with a batch size of 64). In both cases, we used Adam with a learning rate of $10^{-4}$. 

\paragraph{Architectures.}

For EMNIST, in the NP, the latent encoder has 5 dense layers before mean aggregation, and 2 after it. The deterministic encoder has 5 dense layers before mean aggregation. The decoder has 4 dense layers. In the ANP, the latent encoder has 3 dense layers before mean aggregation and 3 after it. The deterministic encoder has 3 dense layers. The decoder has 4 dense layers. The deterministic encoder has 3 dense layers followed by a self-attention layer with 8 heads, then 3 dense layers to transform the keys and queries. The final cross-attention has 8 heads. 
All dense layers are of size 128. 

For CelebA, in the NP, the latent encoder has 6 dense layers before mean aggregation, and 3 after it. The deterministic encoder has 6 dense layers before mean aggregation. The decoder has 5 dense layers. In the ANP, the latent encoder has 4 dense layers before mean aggregation and 3 after it. The deterministic encoder has 4 dense layers. The decoder has 4 dense layers. The deterministic encoder has 4 dense layers followed by a self-attention layer with 8 heads, then 3 dense layers to transform the keys and queries. The final cross-attention has 8 heads. 
All dense layers are of size 128.

\subsection{Electricity forecasting}

\paragraph{Training and testing.} All models were trained for 40,000 iterations with a batch size of 32. We used Adam with a learning rate of $10^{-4}$. 
\section{Further results}
\label{appendix:further_results}

\begin{figure}[h!]
\vskip 0.2in
\begin{center}
\centerline{\includegraphics[width=\columnwidth]{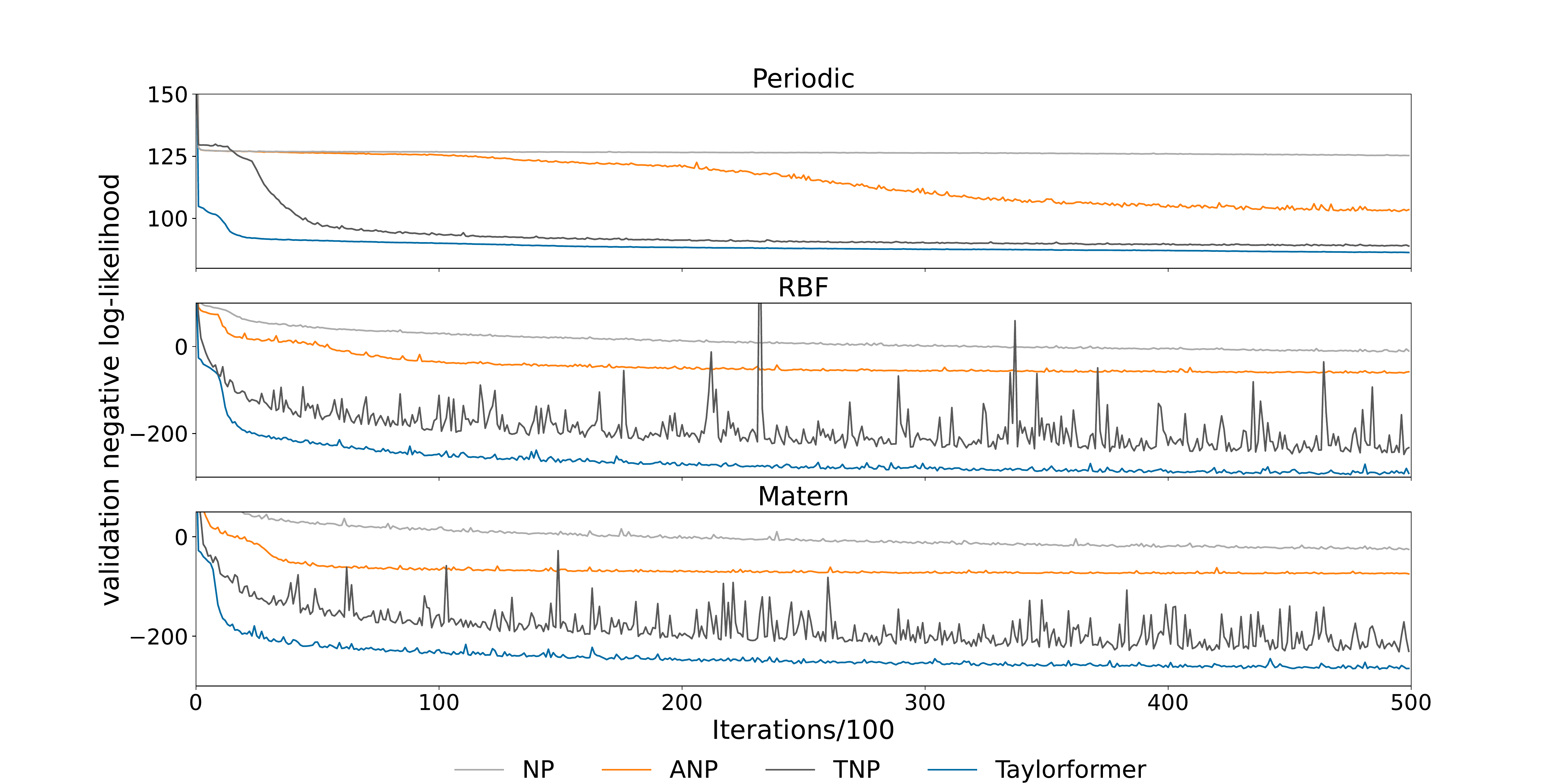}}
\caption{Validation set negative log-likelihood (NLL). Lower is better. Our Taylorformer outperforms NP\cite{Garnelo2018NeuralP}, ANP \cite{Kim2019AttentiveNP}  and TNP \cite{pmlr-v162-nguyen22b} on the meta-learning 1D regression task (see task details in the main text).}
\label{figure: loss_1d_reg}
\end{center}
\vskip -0.2in
\end{figure}

\paragraph{Validation loss for 1D regression.} For all the 1D regression experiments with GPs, we see in Figure \ref{figure: loss_1d_reg} that the validation set negative log-likelihood is lowest for the Taylorformer compared to other Neural Process models. The TNP has a particularly noisy loss curve.

\paragraph{Log-likelihood results for forecasting experiments.}

Please refer to table \ref{table:forecasting_likelihood} for results.

\begin{table*}[tp]%
\caption{Negative log-likelihood on a test set. Lower is better. Taylorformer outperforms three state-of-the-art models on three forecasting tasks (details in text). Each task is trained and evaluated with different prediction lengths $n_{T} \in \{96, 192, 336, 720\}$ given a fixed 96-context length. Each model is run five times to report means with standard deviation results.}
\centering% 
\small
\begin{tabular}{llcccc}
\toprule%
% \multicolumn{2}{l}{Models}  
&&&Negative Log-Likelihood\\
& \multicolumn{2}{c}{\textbf{Taylorformer}}   &  \multicolumn{1}{c}{TNP} & \multicolumn{1}{c}{Autoformer} & \multicolumn{1}{c}{Informer}\\ 
\otoprule%
\multicolumn{1}{|l|}{\multirow{4}{*}{\rotatebox[origin=r]{90}{ETT}}} & 
                            96  &  \boldmath{$-2.69 \pm 0.02$} &$-2.47 \pm 0.05$ & $0.65  \pm 0.16$&
                              $0.61  \pm 0.16$
                              \\ \multicolumn{1}{|l|}{} 
                              & 192 & \boldmath{$-2.68 \pm 0.01$} & $-2.42 \pm 0.14$&$0.87 \pm 0.13$
                              &$0.92  \pm 0.14$\\ \multicolumn{1}{|l|}{} 
                              & 336 & \boldmath{$-2.64 \pm 0.01$} & $-2.48 \pm 0.02$ & $0.88\pm 0.15$&
                              $0.87  \pm 0.14$
                               \\ \multicolumn{1}{|l|}{} 
                              & 720 & \boldmath{$-2.56 \pm 0.01$} &$-2.38 \pm 0.03$ & $1.38\pm 0.07$& $1.01 \pm 0.13$\\\midrule
\multicolumn{1}{|l|}{\multirow{4}{*}{\rotatebox[origin=r]{90}{Exchange}}}& 96  & \boldmath{$-1.31 \pm 0.04$} & $0.27 \pm 1.11$& $0.65 \pm 0.15$& 
$0.86\pm0.11$ \\
                            \multicolumn{1}{|l|}{}   & 192 & \boldmath{$-1.21 \pm 0.03$} & $0.38 \pm 0.40$ & $0.87 \pm 0.13$& $1.33\pm 0.15$\\
                             \multicolumn{1}{|l|}{}  & 336 & \boldmath{$-1.05 \pm 0.04$} & $1.21 \pm 0.47$ &$1.19  \pm 0.09$& $1.80 \pm 0.07$\\
                           \multicolumn{1}{|l|}{}    & 720 & \boldmath{$-0.68 \pm 0.02$}  & $0.94 \pm 0.55$ &$1.05 \pm 0.13$& $1.52 \pm 0.06$\\[1ex]\midrule 
\multicolumn{1}{|l|}{\multirow{4}{*}{\rotatebox[origin=r]{90}{Electricity}}}

                              & 96    
                              & \boldmath{$-0.53 \pm 0.01$}  
                              
                              &
                              $-0.42 \pm 0.03$
                              &
                              
                              $0.38 \pm 0.13$
                              &
                              
                              $0.41 \pm 0.02$                               
                              \\  \multicolumn{1}{|l|}{} 
                              & 192 & 
                             \boldmath{$-0.50 \pm 0.00$}
                
                              &
                              $-0.40 \pm 0.02$
                 
                              & 
                              $0.35 \pm 0.01$
             
                              &
                              $0.46 \pm 0.04$ 
                               
                              \\ \multicolumn{1}{|l|}{} 

                              & 336 &  
                              \boldmath{$-0.41 \pm 0.03$}
                              & $-0.28 \pm 0.08$ 
                              & $0.44 \pm 0.04$
                              & $0.47 \pm 0.05$
                              \\ \multicolumn{1}{|l|}{} 

                              & 720 & 
                              \boldmath{$-0.44 \pm 0.03$}  
                              &$-0.25 \pm 0.13$
                              &$0.64 \pm 0.09$
                              &$0.39 \pm 0.05$ 
                              \\[2ex]

\bottomrule 
\end{tabular}

\label{table:forecasting_likelihood}
\end{table*}

\begin{figure}[!h]
% \vskip 0.2in
\begin{center}
\centerline{\includegraphics[width=0.5\textwidth]{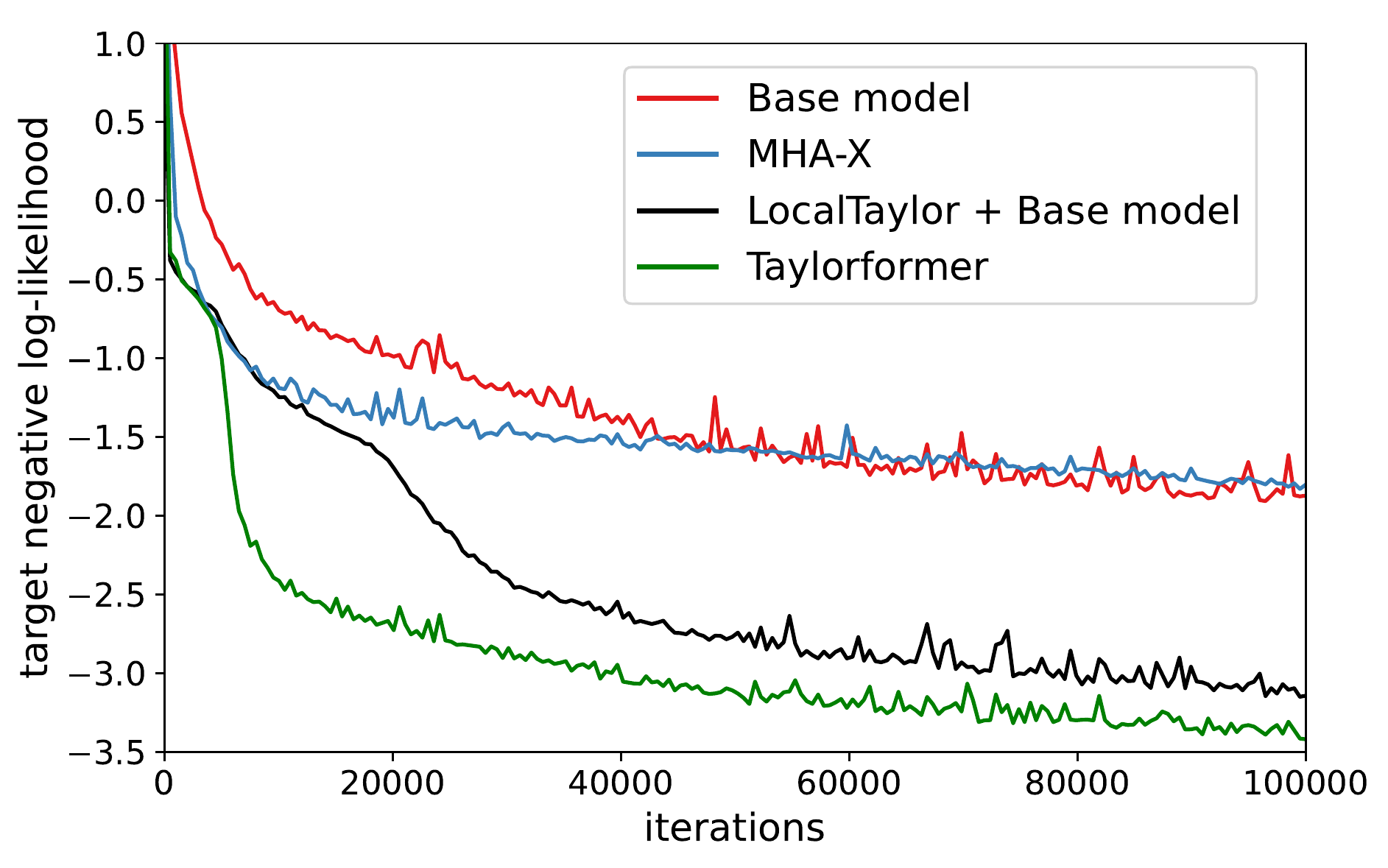}}
\caption{Ablations for our model showing that using both the LocalTaylor wrapper and the MHA-X together (green line) contributes to the improved results. This is shown for a 1D regression task (GP RBF kernel).}
\label{figure:ablation}
\end{center}
\vskip -0.2in
\end{figure}

\paragraph{Ablation study.} The results of our ablation study are shown in Figure \ref{figure:ablation}. We see that it is the combination of both the LocalTaylor approach and the MHA-X attention block which yields the best outcome.

% Figure \ref{figure:val_loss} shows the validation log-likelihood curves for the TNP and ATP across five tasks (GP regression and image completion). The EMNIST curves are far smoother as both models have a lower bound on the standard deviation (as done in \cite{pmlr-v162-nguyen22b}). 

% \begin{figure*}[ht]
% \vskip 0.2in
% \begin{center}
% \centerline{\includegraphics[scale=0.75]{figures/val_losses_plot.pdf}}
% \caption{Validation log-likelihood curves for five of our experiments, for the TNP and ATP. The ATP curves tend to be smoother.}
% \label{figure:val_loss}
% \end{center}
% \vskip -0.2in
% \end{figure*}

\paragraph{Electricity forecasting samples.}

We show samples from the Taylorformer and the TNP for the electricity forecasting task for target-336-context-96. Both models do well but the Taylorformer tracks the peaks and troughs better.

\begin{figure}[ht]
\vskip 0.2in
\begin{center}
\centerline{\includegraphics[width=0.8\columnwidth]{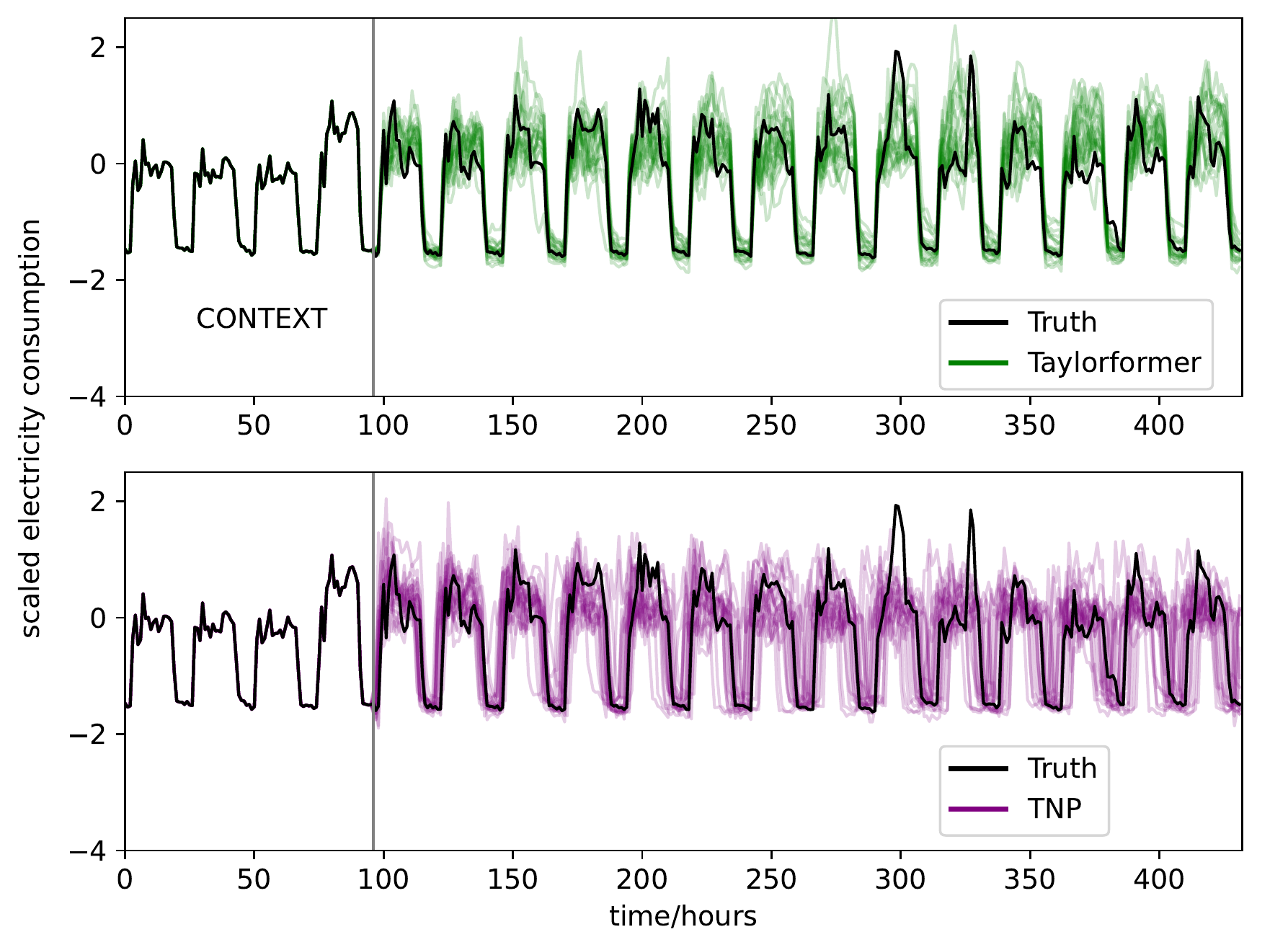}}
\caption{20 samples from our model and the TNP on the electricity task. Ours tracks the periodic trend remarkably well. Although the TNP tracks the periodic trend, it struggles to maintain the appropriate periodicity, especially at longer lead times. The Taylorformer appears to have better-calibrated uncertainty, too: it is more uncertain about the peaks in consumption, and so can produce time-series samples that can capture the large peak around hour 300, unlike the TNP.}
\label{figure:electricity_samples}
\end{center}
\vskip -0.2in
\end{figure}
\section{Code}

Our code can be found at \url{https://github.com/oremnirv/Taylorformer}.
\section{Model implementation} \label{full_imp}

\subsection{Embedding} 
\label{embed}
We follow a similar formulation to the embedding from \citet{Vaswani2017}, but adapt it to continuous values.
\begin{align}
    & PE(t, 2i) = \sin(\dfrac{t/\delta_t}{(tmax/\delta_t)^{2i/d}})\\
    & PE(t, 2i+1) = \cos(\dfrac{t/\delta_t}{(tmax/\delta_t)^{2i/d}})
\end{align}
where $d$ is the total embedding dimension and $i$ runs from $1$ to $d$. We will denote the vector of all embedding dimensions for a specific $t$ as $\operatorname{PE}(t)$.

\subsection{Model implementation} 
We present a generic implementation of our training for batch data in algorithm \ref{alg:training}. The algorithm notation follows the notation presented in the main text.
We denote the mask as $M$. It follows the same masking described in the main text and in Figure \ref{fig:architecture}. The $\operatorname{PositionalEncoding}$ is described in \ref{embed}. By $\operatorname{Split}(\mathbf{W}$) we refer to the operation of extracting subsets of variables from $\mathbf{W}$.  

% We want to present a generic application of our algorithm for batch data with $B$ sequences. $[X_{C}, X_{T}], [Y_{C}, Y_{T}]   \in \mathbf{R}^{B\times \ell \times 1}$ so we will introduce a bit more notation for clarity. By a bold capital letter we refer to a vector. So if in the main text we introduced, for example, the term $x'_i$, then by $\textbf{X}'$ we refer to the sequence $[\{x'_{1}, \dots, x'_{\ell}\}]$. By $\textbf{a}_{B \times \ell \times K}$ we mean a matrix of size $B\times \ell \times K$ all filled with the same value $a$, by \text{MLP}($\cdot$) we mean a generic neural network function, by $\partial$ we refer to the feature extraction algorithm in \ref{alg: FE}. The \text{Add} operator, \text{LayerNorm} and \text{FFN} are all the same as in \cite{Vaswani2017}.  By Mask($\cdot, n_C$) we refer to the operation of masking the derivative info and the actual value of the $y_i$ for all $i$ in the target set of the sequence. We also declare the varibales $x''_i = (x_{n(i)},\, x_i-x_{n(i)})$ and $y'''_i = (y_{n(i)},y_i - y_{n(i)}, d_i)$. By $\textbf{U}_{-1}$ we refer to the last element of a vector $\textbf{U}$ and by $\textbf{U}_{i:j}$ we refer to elements $i$ to $j-1$ of a vector $\textbf{U}$. 
% The full implementation is given in algorithm \ref{alg:training}. In \ref{alg:inference} we give implementation for the inference stage. 

\begin{algorithm}[H]
   \caption{Model implementation for batch data -- training phase}
   \label{alg:training}
\begin{algorithmic}
   \STATE {\bfseries Input:} data $\textbf{X} = [\textbf{X}_{C}, \textbf{X}_{T}]$, $\textbf{X} \in \mathbf{R}^{B \times \ell \times 1}$, $\textbf{Y} =[\textbf{Y}_{C},\textbf{Y}_{T}]$, 
   $\textbf{Y} \in \mathbf{R}^{B\times \ell \times \alpha}$, mask $M$; batch size $B$; context length $n_C$; target length $n_T$; sequence length $\ell$; embedding dimension $d$; output dimension $d_O$; \text{MHA} layers $N$.
   \STATE
   \STATE  $\textbf{X}' = \operatorname{PositionalEncoding}(\textbf{X}$) \hfill\COMMENT{$\textbf{X}' \in \mathbf{R}^{B \times \ell \times d}$ }
   \STATE
   % X, Y, X'

   \STATE $\mathbf{W} = \operatorname{Concat}[\text{i for i in} \operatorname{RepresentationExtractor}(X_T^i, \mathbf{X}_C, \mathbf{Y}_C;q)]$

   % $0th order is x and y $
   % $1st order is x, y, dx, dy, dx/dy and dx/dy_closest$
   % $2nd order is the above, and d2x,d2y,d2x/d2y, and d2x/d2y closest$
   
   \hfill\COMMENT{$\mathbf{W} \in \mathbf{R}^{B\times \ell \times (d + \alpha + 4q) }$}
   \STATE
    \STATE
    $\textbf{X}^{\text{fe}}$, $\textbf{Y}^{\text{fe}}$, 
   $\textbf{Y}^{\text{seen}} = \operatorname{Split}(\mathbf{W})$
    \hfill\COMMENT{$\textbf{X}^{\text{fe}} \in \mathbf{R}^{B\times \ell \times (d+2q)}$,  $\textbf{Y}^{\text{fe}} \in \mathbf{R}^{B\times \ell \times (\alpha+2\alpha q)}$, $\textbf{Y}^{\text{seen}} \in \mathbf{R}^{B\times \ell \times (2\alpha q)}$}
   \STATE
   \STATE $\textbf{Q}{_C}^{\text{MHA-XY}} = 
   [\textbf{X}_C^{\text{fe}}, \textbf{Y}_C^{\text{fe}}$, 
   $\textbf{Y}_C^{\text{seen}}$,
   1]   \hfill\COMMENT{$\textbf{Q}{_C}^{\text{MHA-XY}} \in \mathbf{R}^{B \times n_C \times (4\alpha q + \alpha + 2q+ d + 1)}$ }
   \STATE
   \STATE $\textbf{Q}{_T}^{\text{MHA-XY}} = 
   [\textbf{X}_T^{\text{fe}}, 0, 
   \textbf{Y}_T^{\text{seen}},
   0]$   \hfill\COMMENT{$\textbf{Q}{_T}^{\text{MHA-XY}} \in \mathbf{R}^{B \times n_T \times (4\alpha q + \alpha + 2q+ d + 1)}$ }
   \STATE
   \STATE
   $\textbf{Q}^{\text{MHA-XY}} = \operatorname{Concat}[\textbf{Q}{_C},\textbf{Q}{_T}]$
   \hfill\COMMENT{$\textbf{Q}^{\text{MHA-XY}} \in \mathbf{R}^{B \times \ell \times (4\alpha q + \alpha + 2q+ d + 1)}$ }
   \STATE
   \STATE    $\textbf{K}^{\text{MHA-XY}} = 
   [\textbf{X}^{\text{fe}}, \textbf{Y}^{\text{fe}}, 
   \textbf{Y}^{\text{seen}},
   1]$
   \hfill
    \COMMENT{$\mathbf{K}^{\text{MHA-XY}}\in \mathbf{R}^{B \times \ell \times (4\alpha q + \alpha + 2q+ d + 1)}$}
   \STATE
   \STATE
   $\textbf{V}^{\text{MHA-XY}} =\textbf{K}^{\text{MHA-XY}}$

    \STATE
    \STATE
    $\textbf{Q}^{\text{MHA-X}} = \textbf{K}^{\text{MHA-X}}
    = \textbf{V}^{\text{MHA-X}} = \mathbf{X}^\text{fe}$
    \STATE

    \STATE
    $O_X = \operatorname{MultiHead}(\mathbf{Q}^{\text{MHA-X}},\mathbf{K}^{\text{MHA-X}},\mathbf{V}^{\text{MHA-X}},\text{mask}=M)$  
    \hfill
    \COMMENT{$O_X \in \mathbf{R}^{B \times \ell \times d_O}$}
    
\STATE
\STATE
     $O_X = \operatorname{LayerNorm}(\operatorname{Add}(O_X,\operatorname{Dense}(\mathbf{Q}^\text{MHA-X})))$
     \hfill
     \COMMENT{$O_X \in \mathbf{R}^{B \times \ell \times d_O}$}

\STATE

\STATE

    $O_X = \operatorname{LayerNorm}(\operatorname{Add}(O_{X},\operatorname{Linear}(\operatorname{Dense}(O_X))))$
      \hfill
     \COMMENT{$O_X \in \mathbf{R}^{B \times \ell \times d_O}$}

\STATE
\STATE
$O_{XY} =\operatorname{MultiHead}(\mathbf{Q}^{\text{MHA-XY}},\mathbf{K}^{\text{MHA-XY}},\mathbf{V}^{\text{MHA-XY}},\text{mask}=M)$
\hfill
\COMMENT{$O_{XY} \in \mathbf{R}^{B \times \ell \times d_O}$}

\STATE
\STATE
  $O_{XY} = \operatorname{LayerNorm}(\operatorname{Add}(O_{XY},\operatorname{Dense}(\mathbf{Q}^\text{MHA-XY})))$
\hfill
\COMMENT{$O_{XY} \in \mathbf{R}^{B \times \ell \times d_O}$}

\STATE
\STATE
    $O_{XY} = \operatorname{LayerNorm}(\operatorname{Add}(O_{XY},\operatorname{Linear}(\operatorname{Dense}(O_{XY}))))$
 \hfill
\COMMENT{$O_{XY} \in \mathbf{R}^{B \times \ell \times d_O}$}   

    \STATE
   \FOR{$j=1$ {\bfseries to} $N-2$}
        \STATE
        \STATE  $\mathbf{Q}^{\text{MHA-X}}= O_X$,  $\mathbf{Q}^{\text{MHA-XY}}= O_{XY}$ 
        \STATE
        \STATE $O_X = \operatorname{MultiHead}(O_X, O_X, O_X, \text{mask}=M)$ \hfill\COMMENT{$O_X \in \mathbf{R}^{B \times \ell \times d_O}$}
        \STATE
        \STATE
        $O_X = \operatorname{LayerNorm}(\operatorname{Add}(O_X,\mathbf{Q}^\text{MHA-X}))$
        \hfill\COMMENT{$O_X \in  \mathbf{R}^{B \times \ell \times d_O}$}
        \STATE
        \STATE
        $O_X = \operatorname{LayerNorm}(\operatorname{Add}(O_X,\operatorname{Linear}(\operatorname{Dense}(O_X)))$
        \hfill\COMMENT{$O_X \in  \mathbf{R}^{B \times \ell \times d_O}$}
        \STATE
        \STATE
        $O_{XY} =\operatorname{MultiHead}(O_{XY},O_{XY},O_{XY},\text{mask}=M) $
        \hfill \COMMENT{$O_{XY} \in \mathbf{R}^{B \times \ell \times d_O}$}
        \STATE
        \STATE
        $O_{XY} = \operatorname{LayerNorm}(\operatorname{Add}(O_{XY},\mathbf{Q}^\text{MHA-XY}))$
        \hfill
        \COMMENT{$O_{XY} \in \mathbf{R}^{B \times \ell \times d_O}$}
        \STATE
        \STATE
        $O_{XY} = \operatorname{LayerNorm}(\operatorname{Add}(O_{XY},\operatorname{Linear}(\operatorname{Dense}(O_{XY})))$
        \hfill
        \COMMENT{$O_{XY} \in \mathbf{R}^{B \times \ell \times d_O}$}
        
        \STATE
    \ENDFOR

 \end{algorithmic}
\end{algorithm}

\begin{algorithm}[H]
\begin{algorithmic}

    \STATE
$\mathbf{Q}^{\text{MHA-X}}= O_X$,  $\mathbf{Q}^{\text{MHA-XY}}= O_{XY}$

\STATE
\STATE
    $O_X = \operatorname{MultiHead}(O_X,O_X,\mathbf{Y},\text{mask}=M)$
    \hfill\COMMENT{$O_X \in \mathbf{R}^{B \times \ell \times d_O}$}
    \STATE
    \STATE
        $O_X = \operatorname{LayerNorm}(\operatorname{Add}(O_X,\mathbf{Q}^\text{MHA-X}))$
        \hfill\COMMENT{$O_X \in \mathbf{R}^{B \times \ell \times d_O}$}
    \STATE
    \STATE
     $O_X = \operatorname{LayerNorm}(\operatorname{Add}(O_X,\operatorname{Linear}(\operatorname{Dense}(O_X)))$
     \hfill\COMMENT{$O_X \in  \mathbf{R}^{B \times \ell \times d_O}$}
     \STATE

    \STATE
    $O_{XY} = \operatorname{MultiHead}(O_{XY},O_{XY},O_{XY},\text{mask}=M)$
    \hfill
        \COMMENT{$O_{XY} \in \mathbf{R}^{B \times \ell \times d_O}$}

\STATE
        \STATE
        $O_{XY} = \operatorname{LayerNorm}(\operatorname{Add}(O_{XY},\mathbf{Q}^\text{MHA-XY}))$
        \hfill
        \COMMENT{$O_{XY} \in \mathbf{R}^{B \times \ell \times d_O}$}
        \STATE
        \STATE
        $O_{XY} = \operatorname{LayerNorm}(\operatorname{Add}(O_{XY},\operatorname{Linear}(\operatorname{Dense}(O_{XY})))$
        \hfill
        \COMMENT{$O_{XY} \in \mathbf{R}^{B \times \ell \times d_O}$}

\STATE
\STATE
    
    $O = \operatorname{Linear}(\operatorname{Concat}(O_X,O_{XY}))$
            \hfill
        \COMMENT{$O\in \mathbf{R}^{B \times \ell \times 2\alpha}$}
    \STATE
    \STATE
    $A,B = O[:,:,:\alpha],O[:,:,\alpha:]$
      \hfill
        \COMMENT{$A, B \in \mathbf{R}^{B \times \ell \times \alpha}$}
    \STATE
    \STATE
    $\mathbf{\mu},\mathbf{\sigma} = A + \mathbf{Y}^{n(I)}, B $
    \hfill
        \COMMENT{$\mathbf{Y}^{n(I)} \subset  \mathbf{W}$,  $\mathbf{\mu},\mathbf{\sigma} \in \mathbf{R}^{B \times \ell \times \alpha}$}

    % \STATE $\Bar{\mu}, \Bar{\Sigma} = MLP(o_{H}) + \textbf{Y}''$ \hfill\COMMENT{ $\textbf{Y}''$ is a feature in $\textbf{Y}'''$} 
    % \STATE
    \STATE {\bfseries Return}  ${\mu}[:, n_C:,:], {\sigma}[:, n_C:,:]$
\end{algorithmic}
\end{algorithm}

% \begin{algorithm}[tb]
%    \caption{Model implementation for one sequence -- inference phase}
%    \label{alg:inference}
% \begin{algorithmic}
%    \STATE {\bfseries Input:} data $ \textbf{X} = [\textbf{X}_{c}$, $\textbf{X}_{T}]$, $\textbf{Y}_{c}$, mask $\textbf{M}$; context length $n_C$; sequence length $\ell$.
%    \STATE
%    \FOR{$j=1$ {\bfseries to} $(\ell - n_C + 1)$}
%    \STATE
%    \STATE $\Bar{\mathbf{\mu}}$, $\Bar{\mathbf{\Sigma}}$ =         \text{Algorithm-2}($\textbf{X}_{0:n_C+j}$, $\textbf{Y}_C$, $\textbf{M}_{0:n_C+j, 0:n_C+j}$) \hfill\COMMENT{$\Bar{\mathbf{\mu}} \in \mathbf{R}^{\ell - n_C}$, $\Bar{\mathbf{\Sigma}} \in \mathbf{R}^{\ell - n_C}$ and 
%    \text{Algorithm-2} refers to \ref{alg:training}}
%    \STATE
%    \STATE $y_{n_C+j}$ = \text{Sample}($\operatorname{Normal}(\Bar{\mathbf{\mu}}[0], \Bar{\mathbf{\Sigma}}[0])$) 
%    \STATE
%    \STATE \textbf{Set} $\textbf{Y}_C = [\textbf{Y}_C, y_{n_C+j}]$
%     \ENDFOR
%     \STATE
%     \STATE {\bfseries Return:}  $\textbf{Y}_C$  \hfill\COMMENT{$\textbf{Y}_C \in \mathbf{R}^{\ell}$}
% \end{algorithmic}
% \end{algorithm}

%%%%%%%%%%%%%%%%%%%%%%%%%%%%%%%%%%%%%%%%%%%%%%%%%%%%%%%%%%%%%%%%%%%%%%%%%%%%%%%
%%%%%%%%%%%%%%%%%%%%%%%%%%%%%%%%%%%%%%%%%%%%%%%%%%%%%%%%%%%%%%%%%%%%%%%%%%%%%%%

\end{document}